\pdfoutput=1

\documentclass[11pt,table,dvipsnames]{article}

\usepackage[final]{acl}

\usepackage{times}
\usepackage{latexsym}

\usepackage[T1]{fontenc}

\usepackage[utf8]{inputenc}

\usepackage{microtype}

\usepackage{inconsolata}

\usepackage{graphicx}

\usepackage{latexsym}
\usepackage{dsfont}
\usepackage{booktabs} 
\usepackage{subfigure}
\usepackage{CJKutf8}
\usepackage{graphicx}
\usepackage{bigstrut}
\usepackage{multirow}
\usepackage{amsmath}
\usepackage{multirow, makecell, caption}
\usepackage{floatrow}
\newfloatcommand{capbtabbox}{table}[][\FBwidth]
\usepackage[normalem]{ulem}
\usepackage{amssymb}
\usepackage{amsfonts}
\usepackage{multicol}
\usepackage{tipa}
\usepackage{amsmath}
\usepackage{bm}
\usepackage[ruled,linesnumbered]{algorithm2e}
\usepackage{tikz}
\usepackage{paralist}
\usepackage{IEEEtrantools}
\usepackage[switch]{lineno}
\usepackage{enumitem}
\usepackage{multirow, makecell, caption}
\usepackage{arydshln}
\usepackage{verbatim}
\usepackage[normalem]{ulem}
\usepackage{amssymb}
\usepackage{amsfonts}
\usepackage{multicol}
\usepackage{amssymb}
\usepackage{pifont}
\usepackage{csquotes}
\usepackage{xspace}
\usepackage{paralist}
\usepackage{mdwlist}
\usepackage{subfigure}
\usepackage{makecell}
\usepackage{array}
\usepackage{bm}
\usepackage{colortbl}
\usepackage{dsfont}
\usepackage{url}
\usepackage{booktabs} 
\usepackage{graphicx}
\usepackage{tabularx}
\usepackage{bbm}
\usepackage{pgfplots}
\usepackage{soul} 
\usepackage{listings}
\usepackage[tableposition=top]{caption}
\usepackage{color,xcolor} 
\usepackage{multirow} 

\usepackage{float} 
\usepackage[section]{placeins}

\usepackage{float} 
\usepackage[section]{placeins}
\usepackage{mdframed}
\usepackage{tcolorbox}
\newcommand{\method}{\textsc{DeePer}\xspace}

\makeatletter
\def\adl@drawiv#1#2#3{%
        \hskip.5\tabcolsep
        \xleaders#3{#2.5\@tempdimb #1{1}#2.5\@tempdimb}%
                #2\z@ plus1fil minus1fil\relax
        \hskip.5\tabcolsep}
\newcommand{\cdashlinelr}[1]{%
  \noalign{\vskip\aboverulesep
           \global\let\@dashdrawstore\adl@draw
           \global\let\adl@draw\adl@drawiv}
  \cdashline{#1}
  \noalign{\global\let\adl@draw\@dashdrawstore
           \vskip\belowrulesep}}
\makeatother

\author{
Aili Chen\textsuperscript{\rm$\spadesuit$},
Chengyu Du\textsuperscript{\rm$\spadesuit$},
Jiangjie Chen\textsuperscript{\rm$\heartsuit$}\thanks{Part of the work done while at Fudan University.},
 Jinghan Xu\textsuperscript{\rm$\clubsuit$},\\
\bf Yikai Zhang\textsuperscript{\rm$\spadesuit$},
Siyu Yuan\textsuperscript{\rm$\clubsuit$},
Zulong Chen\textsuperscript{\rm$\diamondsuit$},
Liangyue Li\textsuperscript{\rm$\diamondsuit$},
Yanghua Xiao\textsuperscript{\rm$\spadesuit$}\thanks{Corresponding authors.}\\
\textsuperscript{\rm$\spadesuit$}Shanghai Key Laboratory of Data Science,  \\ College of Computer Science and Artificial Intelligence,  Fudan University\quad \\
\textsuperscript{\rm$\heartsuit$}ByteDance Seed\quad 
\textsuperscript{\rm$\clubsuit$}School of Data Science, Fudan University\quad 
\textsuperscript{\rm$\diamondsuit$}Alibaba Group.
\\
\normalsize
\texttt{\{alchen20,shawyh\}@fudan.edu.cn}\quad    
\texttt{\{cydu24,jhxu21,ykzhang22,syyuan21\}@m.fudan.edu.cn}\quad 
\\ 
\normalsize
\texttt{jiangjiec@bytedance.com}\quad
\texttt{\{zulong.czl, liangyue.lly\}@alibaba-inc.com}\quad
}

\title{
\method Insight into Your User: Directed Persona Refinement for \\Dynamic Persona Modeling}

\begin{document}
\sloppy  
\maketitle

\begin{abstract}

To advance personalized applications such as recommendation systems and user behavior prediction, recent research increasingly adopts large language models (LLMs) for human-readable persona modeling. 
In dynamic real-world scenarios, effective persona modeling necessitates leveraging streaming behavior data to continually optimize user personas.
However, existing methods—whether regenerating personas or incrementally extending them with new behaviors—often fail to achieve sustained improvements in persona quality or future behavior prediction accuracy.
To address this, we propose \method, a novel approach for dynamic persona modeling that enables continual persona optimization.
Specifically, we enhance the model’s direction-search capability through an iterative offline reinforcement learning framework, allowing it to automatically identify effective update directions and optimize personas using discrepancies between user behaviors and model predictions.
Extensive experiments on dynamic persona modeling involving 4,800 users across 10 domains highlight \method’s superior persona optimization capabilities, delivering an impressive 32.2\% average reduction in user behavior prediction error over four update rounds—outperforming the best baseline by a remarkable 22.92\%.\footnote{Resources are available at \url{https://github.com/sheep333c/DEEPER.git}.}

\end{abstract}

\section{Introduction}
\label{sec:intro}

Recent studies increasingly utilize Large Language Models (LLMs) \cite{achiam2023gpt, anthropic2024claude3, llama3modelcard} for human-readable and interpretable persona modeling, advancing personalized applications like recommendation and behavior prediction. However, most research focuses on generating personas from static historical data, which fail to capture dynamic behaviors and evolving preferences in real-world interactive scenarios \cite{ wang2023zero, zhou2024language}. This underscores the need for dynamic persona modeling—a pivotal yet underexplored approach that iteratively updates personas using streaming user behavior data to continually enhance their quality.

\begin{figure}[t]
    \centering
    \includegraphics[width=\linewidth]{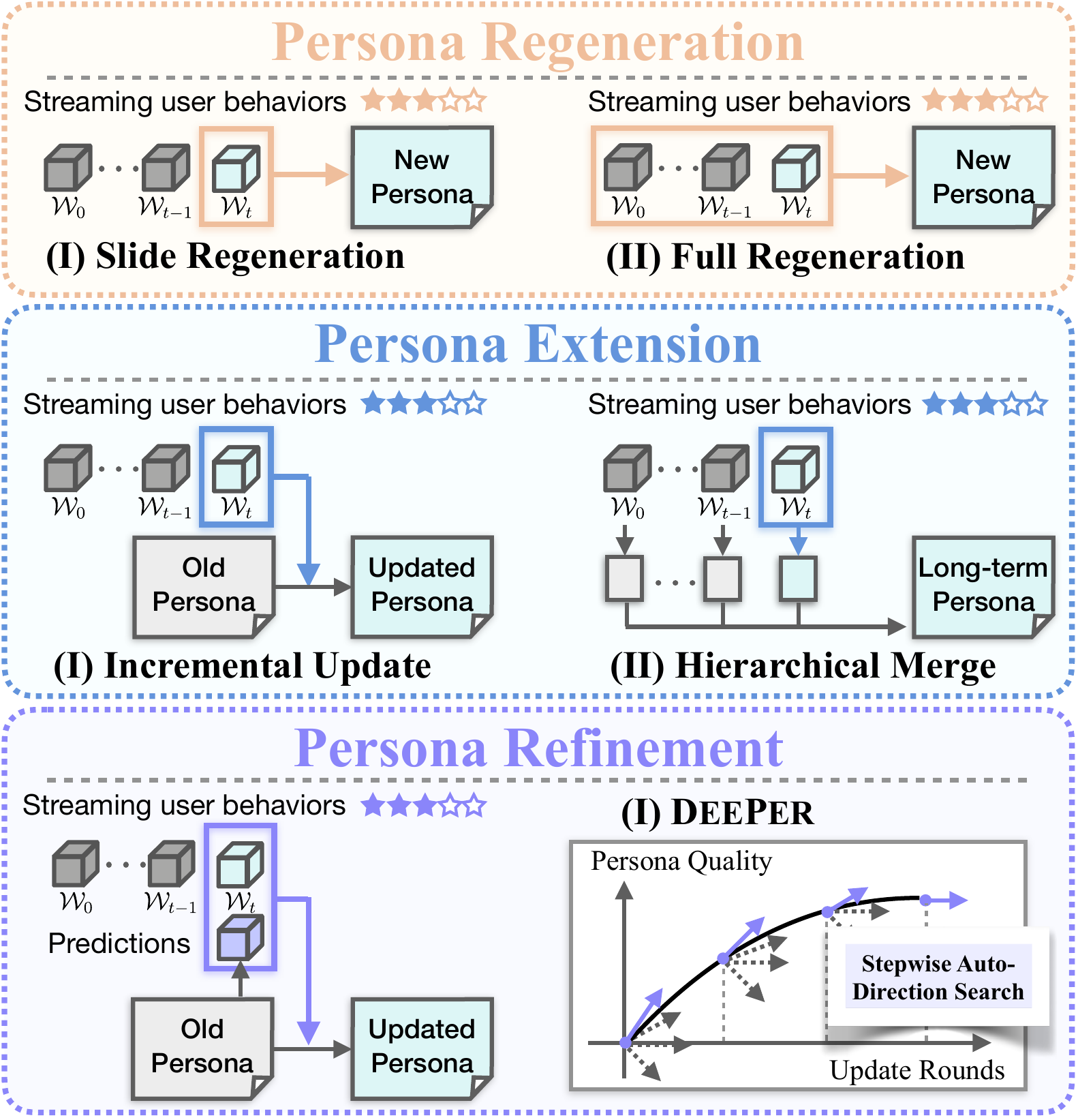}
    \caption{Comparison of dynamic persona modeling paradigms: Regeneration replaces personas, and Extension adds to them, but neither ensures optimization. Our \method, based on Refinement paradigm, uses discrepancies between user behavior and model predictions to identify update directions for continuous optimization.}
    \label{fig:paradigm}
\end{figure}

Existing dynamic persona modeling methods can be broadly categorized into two paradigms: 
(1) \textit{Persona Regeneration}, which updates through complete replacement, rebuilds personas from scratch based on new user behaviors, either by aggregating historical and recent behaviors or by using sliding-window methods \cite{zhou2024language, wu2024rlpf, yang2023palr, xi2024towards}. 
(2) \textit{Persona Extension}, which updates through additive extension, incorporates new user behaviors into existing personas, either by directly integrating them or by merging short-term and long-term personas. \cite{lian2022incremental,liu2024once, tang2023enhancing,yin2023heterogeneous,yuan2024evaluating}.  
However, while these methods enable dynamic updates, they fail to ensure meaningful optimization due to the lack of mechanisms to evaluate update effectiveness and explicitly model the update process. Without validating whether updates enhance persona quality or predictive accuracy, both paradigms risk propagating errors and degrading performance. This highlights a critical challenge in dynamic persona modeling: \textit{Bridging the gap between updating personas and truly optimizing them}.

To better characterize the update process and bridge this gap, we introduce the concept of \textit{update direction}, which uniquely identifies the transformation from an existing persona to an updated one under given signals. It directly determines whether the update improves, degrades, or maintains persona quality within a specific context, serving as a core factor in persona optimization.

However, identifying an effective update direction is challenging due to the fundamental misalignment between the dense natural language persona space and the discrete user behavior space (e.g., ratings):
(1) Behavior signals are insufficient, e.g., a user’s 1-star movie rating does not clearly indicate whether the dissatisfaction is due to the story, pacing, or genre, making it difficult to identify specific errors in the persona.
(2) Evaluating update directions is inherently complex, e.g., even if we adjust the persona to emphasize “plot complexity” or “character development,” it’s unclear which change would lead to better predictions.

To address the challenges, we propose \textbf{\method} (\text{\textbf{D}ir\textbf{e}ct\textbf{e}d \textbf{Pe}rsona \textbf{R}efinement}), a novel approach for LLM-based dynamic persona modeling. 
Specifically, we introduce a new paradigm, \textit{Persona Refinement} (Figure~\ref{fig:paradigm}), which uses discrepancies between user behaviors and model predictions as stronger update signals to expose deficiencies in personas.
To identify effective update directions, we decompose the optimization objective into three direction search goals: \textit{Previous Preservation}, \textit{Current Reflection}, and \textit{Future Advancement}, ensuring stability, adaptability, and task alignment. 
Based on these goals, we design reward functions for clear and measurable assessments of update directions by comparing predictive errors before and after updates.
Finally, we propose an iterative offline reinforcement learning (RL) framework with two training stages, leveraging self-sampling and DPO fine-tuning to progressively enhance the model’s direction search and persona refinement capabilities, ultimately improving prediction accuracy.

Extensive experiments on over 4800 users across 10 domains demonstrate \method’s strong persona optimization and direction search capability.

In summary, our contributions are as follows:

\begin{itemize}[noitemsep,left=0pt]
    \item We identify key limitations in current LLM-based dynamic persona modeling methods, emphasizing the critical gap between persona updating and optimization caused by weak update signals and unclear update direction.

    \item We propose \method, a novel approach to dynamic persona modeling that achieves continual optimization through discrepancy-based update signals and robust direction search.

    \item Extensive experiments demonstrate that \method successfully bridges this gap, outperforming existing methods in dynamic persona modeling.
\end{itemize}

\section{Dynamic Persona Modeling}
\label{sec:preliminary}

Building on prior work\cite{yang2023palr,kang2023llms,zhou2024language}, we formalize the concept of persona quality and the objective of dynamic persona modeling as follows:

\noindent
\paragraph{Definition 1} (\textit{Persona Quality})
The extent to which a persona accurately represents a user’s preferences and behaviors, indicating its ability to predict future behaviors within a specific domain.

\noindent
\paragraph{Definition 2.} (\textit{Persona Optimization})
The updated persona better represents a user than the previous persona, with improved predictive capability within a specific domain.

\noindent
\paragraph{Objective:} (\textit{Continual Persona Optimization})
Iteratively enhance persona quality through multi-round updates, progressively enhancing its predictive capability within a specific domain.

\subsection{Task Formulation}
Consider a user $\mathcal{U}$ in domain $\mathcal{X}$. 
To capture temporal dynamics of user behaviors, we segment user’s online interactions into sequential, time-ordered windows $\mathbf{W}=\{\mathcal{W}_t\}_{t=0}^\mathcal{T}$. 
Each window $\mathcal{W}_t$ contains $N$ interactions, represented by an item list $\mathbf{I}_t = \{i_{t}^{j}\}_{j=1}^N$ and the corresponding user behaviors $\mathbf{O}_t = \{o_{t}^{j}\}_{j=1}^N$. 
As new data arrives at time $t$, the current window $\mathcal{W}_t$ captures interactions from the present period, while $\mathcal{W}_{t-1}$ reflects previous behaviors, and $\mathcal{W}_{t+1}$ outlines future interactions.

The LLM-based dynamic persona modeling pipeline consists of three stages:
\begin{itemize}[noitemsep,left=0pt]
    \item \textbf{Persona Initialization:} At time step $t=0$, the persona $\mathcal{S}_0$ is initialized based on the user behaviors in the initial window $\mathcal{W}_0$.
    \item \textbf{Behavior Observation and Prediction:} In each window $\mathcal{W}_t$, previous persona $\mathcal{S}_{t-1}$ is used to predict user behaviors $\hat{\mathbf{O}}_{t|\mathcal{S}_{t-1}} = \mathcal{P}(\mathcal{S}_{t-1})$, while actual behaviors $\mathbf{O}_t$ are observed.
    \item \textbf{Persona Update:} At the end of each window $\mathcal{W}_t$, the persona updates using new observations.
\end{itemize}

For the first two stages, we use frozen LLM to generate initial personas and predictions across all modeling paradigms. 
The Persona Update stage, however, varies by paradigm and is formulated as:
\begin{itemize}[noitemsep,left=0pt]
\item \textbf{Persona Regeneration:} Rebuild persona at the end of each window $\mathcal{W}_t$ using new behaviors $\mathbf{O}_t$:
\vspace{-2mm}
\[    
    \mathcal{S}_t = f_{\text{regen}}(\mathbf{O}_t). \tag{1} 
    \vspace{-2mm}
\]
\item \textbf{Persona Extension:} Extend the previous persona $\mathcal{S}_{t-1}$ with new behaviors $\mathbf{O}_t$:  
\vspace{-2mm}
\[
\mathcal{S}_t = f_{\text{exten}}(\mathcal{S}_{t-1}, \mathbf{O}_t). \tag{2} 
    \vspace{-2mm}
\]
\item \textbf{Persona Refinement (proposed):} Refine the previous persona $\mathcal{S}_{t-1}$ with new user behaviors $\mathbf{O}_t$, and predicted results 
$\hat{\mathbf{O}}_{t|\mathcal{S}_{t-1}}$:  
\vspace{-2mm}
\[
\mathcal{S}_t = f_{\text{refine}}(\mathcal{S}_{t-1}, \mathbf{O}_t, \hat{\mathbf{O}}_{t|\mathcal{S}_{t-1}}). \tag{3} 
\]
\end{itemize}
\subsection{Task Evaluation}

In this work, we assess persona quality indirectly through performance in a user- and domain-specific task: future behavior prediction. Prediction error, quantified by the Mean Absolute Error (MAE), serves as an indicator of \textit{Persona Quality}:

\vspace{-2mm}
\[
\varepsilon_{t+1|\mathcal{S}_t} = \frac{1}{n} \sum_{j=1}^N \big| \hat{o}_{t+1|\mathcal{S}_t}^{j} - o_{t+1}^{j} \big|. \tag{4}
\vspace{-2mm}
\]
$\mathbf{O}_{t+1} = \{o_{t+1}^j\}_{j=1}^N$ represents user actual behaviors in $\mathcal{W}_{t+1}$, while $\hat{\mathbf{O}}_{t+1|\mathcal{S}_t} = \{\hat{o}_{t+1|\mathcal{S}_t}^j\}_{j=1}^N$ denotes predictions with persona $\mathcal{S}_t$. 

Lower error indicates better alignment. \textit{Persona Optimization} is realized when an updated persona reduces the prediction error for future behaviors:
\vspace{-2mm}
\[
\varepsilon_{t+1|\mathcal{S}_t} < \varepsilon_{t|\mathcal{S}_{t-1}}. \tag{5}
\vspace{-2mm}
\]
Thus, the evaluation of a dynamic persona modeling method is determined by its ability to achieve the objective of \textit{Continual Persona Optimization}, with an effective update strategy evidenced by a progressive reduction in prediction error over time.

\section{\method}
\label{sec:method}

\begin{figure*}[t]
    \centering
    \includegraphics[width=\linewidth]{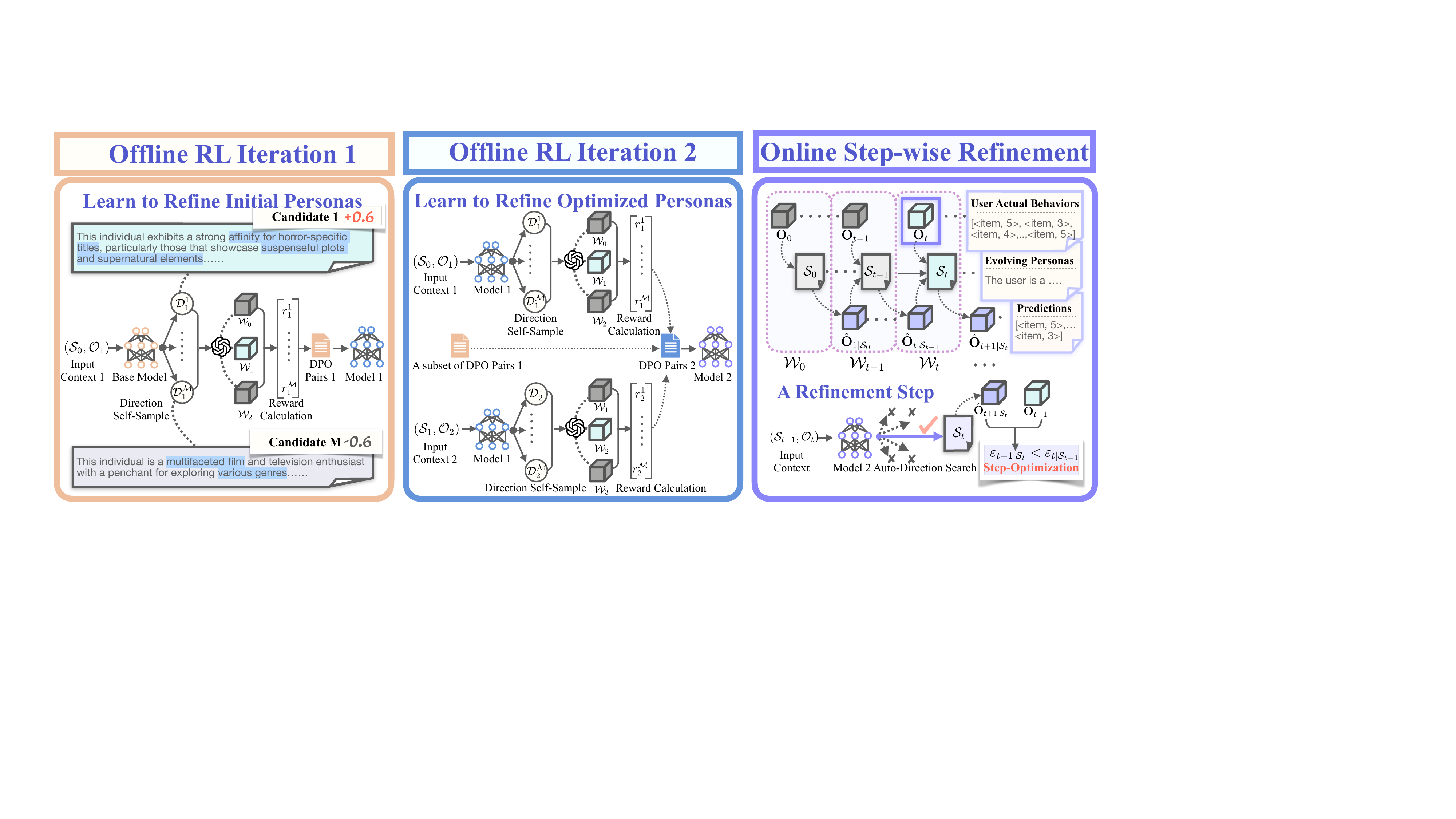}
    \caption{Framework of \method. Grounded in three high-level goals for direction search, the iterative RL framework progressively enhances the model’s refinement capability through two rounds of self-sampling and training. Applied online in multi-round updates, it enables step-wise persona optimization via directed refinement.}
    \label{fig:deeper}
\end{figure*}

Existing regeneration- and extension-based methods enable dynamic updates but fall short in consistent quality improvement, resulting in a misalignment between the \textit{update step} and the \textit{optimization objective}. To address this, we highlight the critical role of update direction in ensuring effective updates and propose the following core proposition:

\noindent  
\textit{Better update directions lead to better personas.}

\noindent 
Instead of directly searching for improved personas, we optimize refinement directions. By incorporating model predictions into the context and defining three high-level goals for direction search, we propose an iterative reinforcement learning framework with a balanced reward function, enabling effective refinement and continual persona optimization.

\subsection{Refinement Step Formulation}

In \method, each persona refinement step at time $t$, can be formulated as a reinforcement learning (RL) task.
The objective is to learn a policy $\pi_\theta$ to identify optimal refinement directions for specific contexts. For a single user $\mathcal{U}$ in domain $\mathcal{X}$, the refinement step can be formulated as:

\begin{itemize}[noitemsep,left=0pt]
\item \textbf{State:} The previous persona $\mathcal{S}_{t-1}$, generated after the $(t-1)$-th refinement round at the end of the previous window $\mathcal{W}_{t-1}$.

\item \textbf{Observation:} The observation at time step $t$, $\mathcal{O}_t = \{\mathbf{O}_t, \hat{\mathbf{O}}_{t|\mathcal{S}_{t-1}}\}$, where $\mathbf{O}_t$ and $\hat{\mathbf{O}}_{t|\mathcal{S}_{t-1}}$ represent the actual and predicted behaviors in the current window $\mathcal{W}_t$.

\item \textbf{Action:} The refined persona $\mathcal{S}_t$, generated after the $t$-th refinement process based on the corresponding $(\mathcal{S}_{t-1}, \mathcal{O}_t)$ of the user. 

\item \textbf{Policy Model:} The refinement model $\pi_\theta$, maps the state and observation to refined persona $\mathcal{S}_t$:
\vspace{-2mm}
\[
\pi_\theta : (\mathcal{S}_{t-1}, \mathcal{O}_t) \to \mathcal{S}_t. \tag{6}
\vspace{-2mm}
\]
\item \textbf{Reward:} The reward $r_t$ quantifies the effectiveness of the refinement process.
\end{itemize}

\subsection{Direction and Goal Definition}
In this work, we formally define the persona refinement direction and its goals as follows:

\noindent
\paragraph{Definition 3.} (\textit{Persona Refinement Direction}) Identify the directed path of a specific persona refinement step, denoted as $\mathcal{D}_t$, which is uniquely determined by the previous persona $\mathcal{S}_{t-1}$, the current observation $\mathcal{O}_t$, and the refined persona $\mathcal{S}_t$:
\vspace{-2mm}
\[
\mathcal{D}_t \leftrightarrow (\mathcal{S}_{t-1}, \mathcal{O}_t; \mathcal{S}_t). \tag{7}
\vspace{-2mm}
\]

We define three high-level goals for direction search, ensuring comprehensive guidance with temporal insights from past, present, and future.

\noindent
\paragraph{Goal 1.} (\textit{Previous Preservation}): 
Retain stable persona traits from historical behaviors to ensure consistency and preserve critical information.

\noindent
\paragraph{Goal 2.} (\textit{Current Reflection}): Adapt to recent user behaviors by incorporating dynamic changes and correcting errors in the previous persona.

\noindent
\paragraph{Goal 3.} (\textit{Future Advancement}): Enhance the persona’s predictive capability for future behaviors.

\subsection{Reward Function Design}
Given the unique correspondence between $\mathcal{D}_t$ and the triplet $(\mathcal{S}_{t-1}, \mathcal{O}_t; \mathcal{S}_t)$, the quality of $\mathcal{D}_t$ directly determines the refined persona’s quality and process effectiveness within context $(\mathcal{S}_{t-1}, \mathcal{O}_t)$. Accordingly, we formalize three goals of \textit{Direction Quality} as reductions in prediction error from refinement across past, current, and future windows, represented by rewards $r_t^{prev}$, $r_t^{curr}$, and $r_t^{fut}$.

\vspace{-2mm}
\[
r_t^{prev} = \varepsilon_{t-1|\mathcal{S}_{t-1}} - \varepsilon_{t-1|\mathcal{S}_t}
\vspace{-2mm}
\]
\[
r_t^{curr} = \varepsilon_{t|\mathcal{S}_{t-1}} - \varepsilon_{t|\mathcal{S}_t}
\vspace{-2mm}
\]
\[
r_t^{fut} = \varepsilon_{t+1|\mathcal{S}_{t-1}} - \varepsilon_{t+1|\mathcal{S}_t}. \tag{8}
\]

\noindent
$\varepsilon_{t-1|\mathcal{S}_{t-1}}$, $\varepsilon_{t|\mathcal{S}_{t-1}}$, and $\varepsilon_{t+1|\mathcal{S}_{t-1}}$ are prediction errors with previous persona $\mathcal{S}_{t-1}$ across $\mathcal{W}_{t-1}$, $\mathcal{W}t$, and $\mathcal{W}_{t+1}$, respectively, while $\varepsilon_{t-1|\mathcal{S}_t}$, $\varepsilon_{t|\mathcal{S}_t}$, and $\varepsilon_{t+1|\mathcal{S}_t}$ are errors with refined persona $\mathcal{S}_t$.

The total reward for a refinement step is:
\vspace{-2mm}
\[
r_t = r_t^{prev}+ r_t^{curr}+r_t^{fut}. \tag{9}
\]

\subsection{Iterative Training Framework}
\method employs an iterative offline RL training framework (Figure~\ref{fig:deeper}): Iteration 1 fine-tunes the base model to refine initial personas (Model 1), while Iteration 2 further enhances it to refine pre-optimized personas (Model 2). Direct Preference Optimization (DPO)\cite{rafailov2024direct} is used to seamlessly integrate rewards into preference pairs, enabling the model to identify better directions through explicit comparisons and supporting scalable iterative fine-tuning.

\paragraph{Iteration 1: Learn to Refine Initial Personas}
Iteration 1 formulates the first refinement step at $t=1$ as an RL task, where we fine-tune the base model to refine initial personas $\mathcal{S}_0$, establishing a baseline policy for direction search and refinement.

\textbf{Context Data Construction}  
First, we initialize personas $\mathcal{S}_0$ for users across multiple domains with their behaviors in window $\mathcal{W}_0$, serving as initial states for refinement processes. The prediction model then predicts user behaviors in $\mathcal{W}_1$ based on $\mathcal{S}_0$. Combining predicted and actual behaviors, we construct observations $\mathcal{O}_1$. Together, $(\mathcal{S}_0, \mathcal{O}_1)$ form the context of the first refinement step.

\textbf{Direction Sampling and Reward Calculation}  
For each context input $(\mathcal{S}_0, \mathcal{O}_1)$, the base model samples $M$ candidate refined personas $\{\mathcal{S}_1^k\}_{k=1}^M$, where each candidate direction $\mathcal{D}_t^k$ is represented by $(\mathcal{S}_0, \mathcal{O}_1; \mathcal{S}_1^k)$. Rewards for these directions as calculated as specified in Equation (9).

\textbf{Preference Pairs Construction and Training}
Refined personas are partitioned into a positive set $\mathcal{S}_1^+$ (rewards $r_t \geq \tau^+$) and a negative set $\mathcal{S}_1^-$ (rewards $r_t \leq \tau^-$) based on reward thresholds. To ensure a clear distinction, we enforce a margin $\delta$, derived from the reward distribution, such that $r_t^w - r_t^l \geq \delta$. The base model is then fine-tuned using DPO with these preference pairs. Following~\cite{gui2024bonbon}, a Supervised Fine-Tuning (SFT) loss is incorporated into the standard DPO objective to maintain alignment with high-quality refinements:
\vspace{-2mm}
\[
\mathcal{L}(\pi_\theta; \pi_{\text{ref}}) = \mathcal{L}_{\text{DPO}}(\pi_\theta; \pi_{\text{ref}}) + \alpha \mathcal{L}_{\text{SFT}}(\pi_\theta). \tag{10}
\]

\paragraph{Iteration 2: Learn to Refine Optimized Personas}
Iteration 2 extends the model's refinement capabilities to handle pre-optimized personas, addressing increased complexity of nuanced refinement tasks.

\textbf{Context Data Construction}
Includes:
(1) Contexts from Iteration 1 $(\mathcal{S}_0, \mathcal{O}_1)$. (2) New contexts $(\mathcal{S}_1, \mathcal{O}_2)$ constructed based on the second refinement step, using $\mathcal{S}_1$ as initial states, with predicted and actual behaviors in $\mathcal{W}_2$ as observations.

\textbf{Direction Sampling and Reward Calculation}
Model 1 is used to sample candidates, following the same procedure of Iteration 1.

\textbf{Preference Pairs Construction and Training}
Similarly to Iteration 1, we construct preference pairs with consistent boundaries for positive and negative sets, with a larger margin $\delta$ to accommodate refined reward distribution and model performance. Model 1 is then fine-tuned with the same combined loss as in Iteration 1, incorporating a subset of preference pairs from Iteration 1 to prevent forgetting and ensure continual learning.

\section{Experiment}
\label{sec:exevalu}

\subsection{Experiment Setup}

\paragraph{Dataset and Task Data Construction}  
We evaluate \method on four real-world datasets across 10 domains, including \textit{MovieLens 20M}\cite{harper2015movielens}, \textit{Food.com Recipes}\cite{majumder2019generating}, \textit{Google Local Reviews}\cite{yan2023personalized,li2022uctopic}, and \textit{Amazon Reviews (2018)}\cite{ni2019justifying}. From six domains, we sample 14,959 users with over 50 ratings (10,800 for training and 4,159 for testing). To assess generalization, an auxiliary test set of 650 users from four unseen domains is constructed. User interactions are segmented into five 10-rating windows, with $\mathcal{W}_0$ used for initial persona generation.

\paragraph{Evaluation} 
As described in Section~\ref{sec:preliminary}, we evaluate the effectiveness of persona update methods based on their ability to achieve \textit{Continual Persona Optimization}, quantified by the reduction in future prediction error $\varepsilon_{t+1|\mathcal{S}_t}$ across update rounds.

\paragraph{Baselines}
We compare against baselines from two paradigms:
1. \textbf{Persona Regeneration}:  
   - \textit{SlideRegen}: Rebuilds personas using only the latest window of behaviors~\cite{yang2023palr}.
   - \textit{FullRegen}: Reconstructs personas by leveraging all historical and recent behaviors~\cite{zhou2024language}.
2. \textbf{Persona Extension}: 
   - \textit{IncUpdate}: Incrementally integrates new behaviors into existing personas~\cite{yuan2024evaluating}. 
   - \textit{HierMerge}\cite{liu2024once}: Hierarchically merges short-term and long-term personas.

\begin{figure}[t]
    \centering    
    \includegraphics[width=\linewidth]{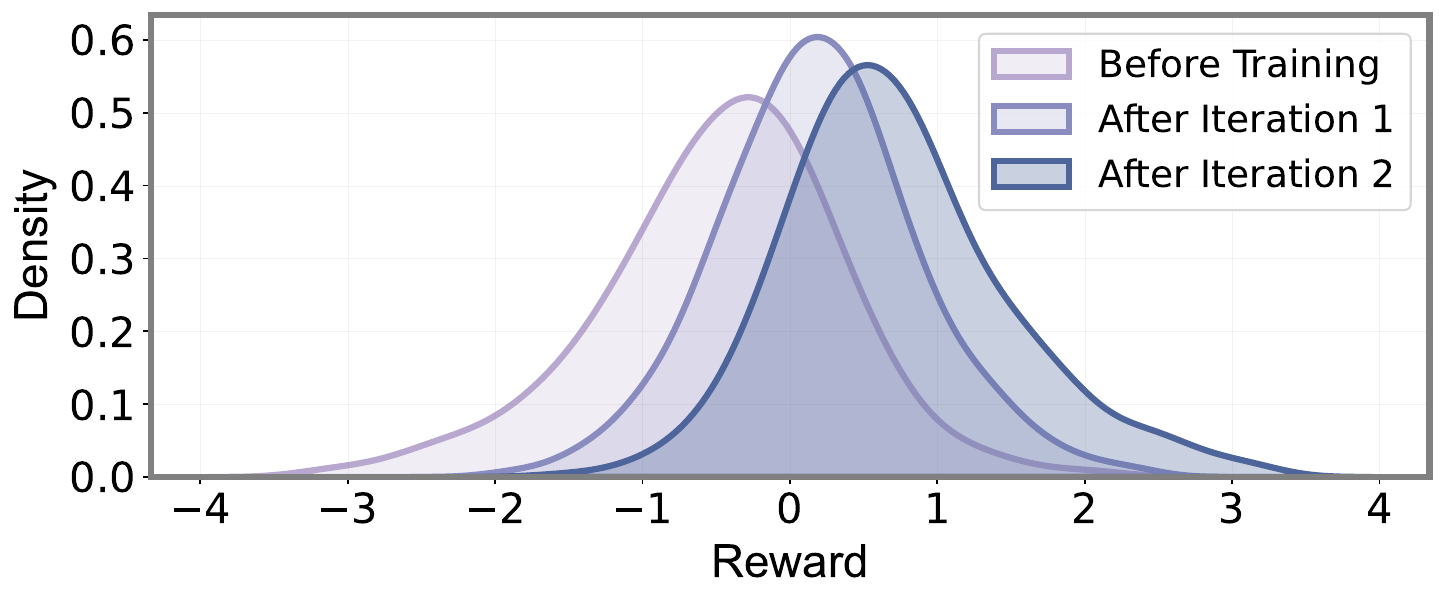}
    \caption{KDE plot illustrating changes in reward distribution across test sets before and after training.}
    \label{fig:reward_kde}
\end{figure}

\paragraph{\method Training Details}  
For \method, we use Llama-3.1-8B-Instruct\cite{ llama3modelcard} as the base policy model, trained iteratively on data from 10,809 users. Each iteration samples 15 candidate personas per input, with reward boundaries $\tau^+ = 0.5$ and $\tau^- = 0$. Iteration 1 applies a margin $\delta = 0.5$, producing 34,782 DPO pairs. Iteration 2 increases $\delta$ to 0.8 and incorporates 5,000 pairs from Iteration 1, resulting in 33,612 pairs. Both iterations use LoRA for fine-tuning with a learning rate of $5 \times 10^{-6}$, 4 training epochs, and a batch size of 128. The SFT loss coefficient ($\alpha$) is set to 0.1. Figure~\ref{fig:reward_kde} illustrates the reward distribution improvements across iterations.

\paragraph{Global and Baseline Settings}

We use the frozen, powerful LLM, GPT-4o-mini\cite{achiam2023gpt}, to generate initial personas and predictions in a zero-shot setting for both training and evaluation, ensuring consistent initial persona quality and unbiased predictions. Additionally, it serves as the backbone for all baselines, offering a robust foundation for comparison.

\begin{figure*}[t]
    \centering
    \includegraphics[width=\linewidth]{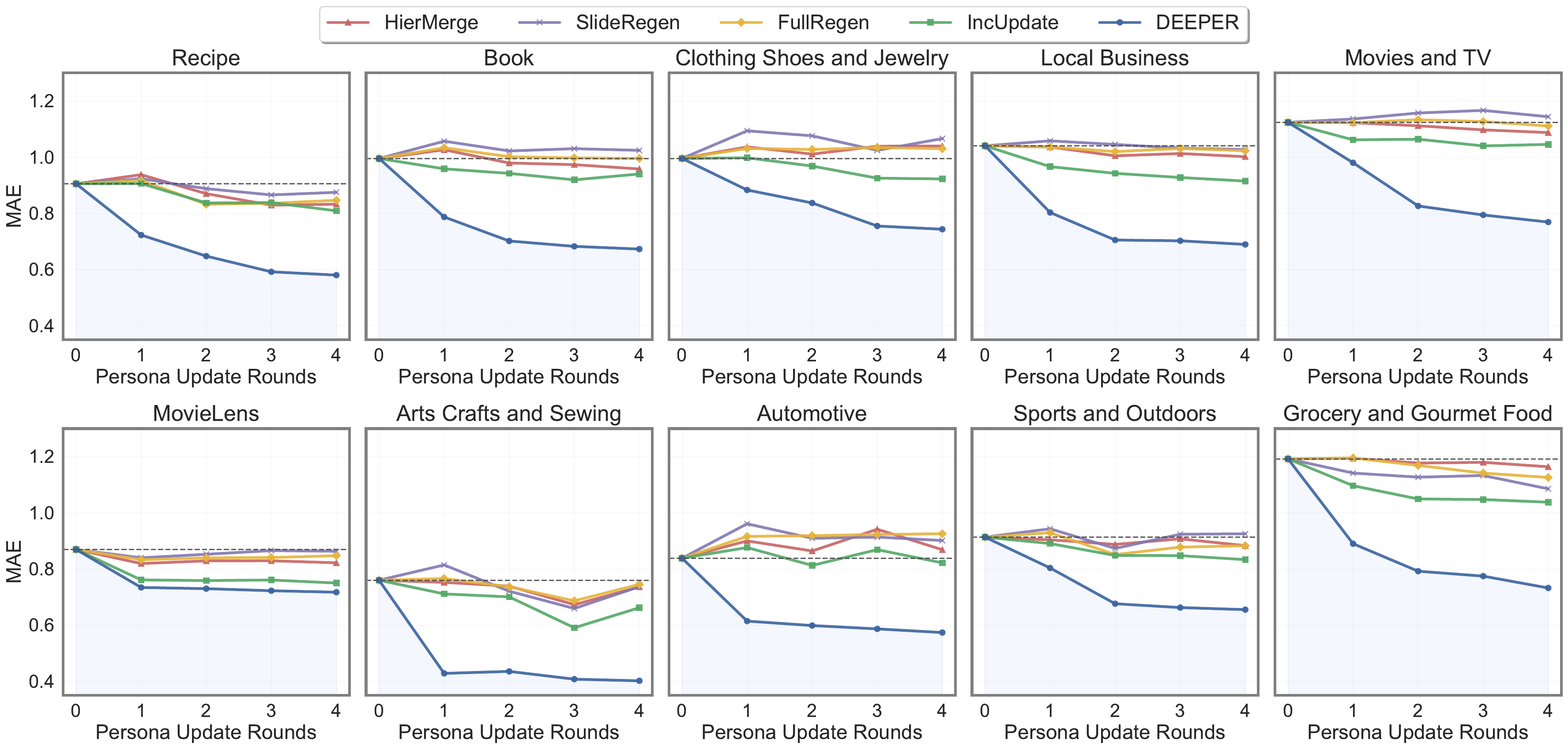}
    \caption{Performance of different methods in dynamic persona modeling over 4 rounds across 10 domains. The first six ((A) \textit{Recipe}, (B) \textit{Book}, (C) \textit{Clothing Shoes and Jewelry}, (D) \textit{Local Business}, (E) \textit{Movies and TV}, (F) \textit{MovieLens}) are seen during training, while ((G) \textit{Arts Crafts and Sewing}, (H) \textit{Automative}, (I) \textit{Sports and Outdoors}, (J) \textit{Grocery and Gourmet Food}) are unseen. In subsequent figures, domains are referred to by their corresponding letters.}
    \label{fig:main_results}
\end{figure*}

\subsection{Main Results}
Figure~\ref{fig:main_results} compares performance of \method and baseline methods over four update rounds across 10 domains in the dynamic persona modeling task.

\paragraph{\method helps continual persona optimization.}  
\method consistently achieves substantial MAE reductions across all 10 domains over four update rounds, with an average decrease of 32.2\%, significantly outperforming extension-based baselines such as \textit{IncUpdate} (9.28\%) and \textit{HierMerge} (3.92\%). Notably, in the unseen domain \textit{Arts Crafts and Sewing}, \method achieves the largest improvement, reducing MAE from 0.76 to 0.40 (47.1\%). In contrast, regeneration-based baselines like \textit{FullRegen} and \textit{SlideRegen} often exhibit minimal or negative gains, highlighting their inability to meet the task objective.

\paragraph{Generalized capability and domain-specific dynamic.}
\method achieves an average MAE reduction of 29.4\% in seen domains and 36.4\% in unseen domains. This emphasizes its generalized optimization capability to diverse and new scenarios. Figure~\ref{fig:main_results} also reveals domain-specific variations in optimization speed, convergence patterns, and improvement potential. For instance, domains like \textit{Automotive} exhibit faster optimization with earlier convergence, while \textit{Movies and TV} shows slower progress and prolonged refinement. These suggest potential influences from varying persona modeling complexities, behavior predictability, and interest stability across domains.

\section{In-depth Analysis}
\label{sec:analysis}

\subsection{What enables \method's effectiveness}

\begin{figure}[t]
    \centering
    \includegraphics[width=\linewidth]
    {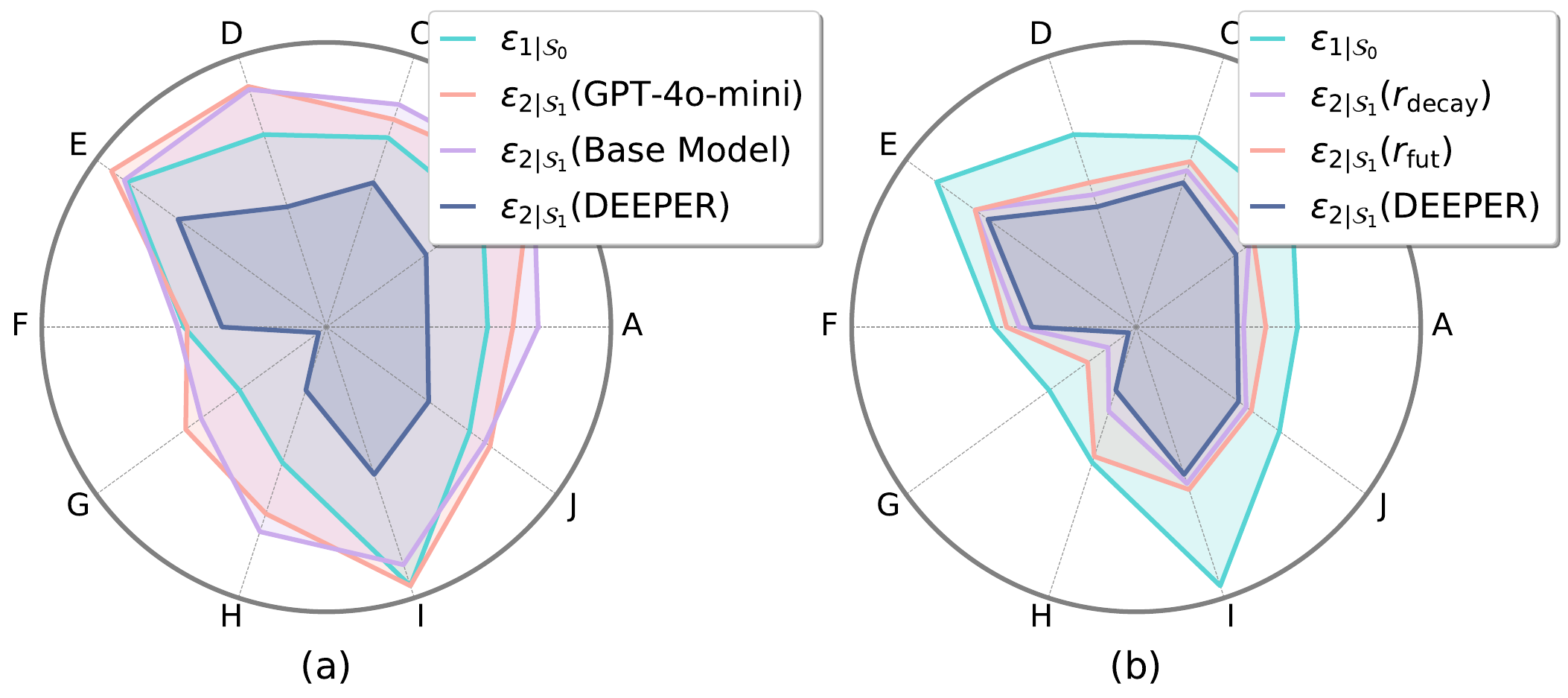}
    \caption{
    (a) Refinement performance of \method compared to frozen models across ten domains, using \(\varepsilon_{1 | \mathcal{S}_0}\) as pre-refinement baseline. 
    (b) Refinement under different reward settings. 
    Smaller areas indicate reduced errors and improved refinement relative to baseline.}
    \label{fig:analysis_radar_part1}
\end{figure}

\paragraph{Direction search enables optimization.}
We first evaluate the necessity of direction search by comparing \method with frozen models (GPT-4o-mini and Llama-3.1-8B-Instruct (Base Model)) which refine personas directly. As shown in Figure~\ref{fig:analysis_radar_part1}(a), both baselines exhibit significant error increases after refinement, underscoring the critical role of direction search in effective optimization.

\vspace{-2mm}

\paragraph{Balanced goals drive better optimization.}
The assessment of Direction Quality is critical to optimization performance. We compare \method’s balanced reward setting(equally weights previous, current, and future goals), with a future-focused reward and a decayed reward (decay factor = 0.5) prioritizing recent goals. As shown in Figure~\ref{fig:analysis_radar_part1}(b), \method consistently outperforms both baselines across all domains. These findings underscore the importance of balanced, goal-driven direction search in enabling effective persona refinement.

\newcolumntype{b}{>{\columncolor{brown!6}}c}
\newcolumntype{q}{>{\columncolor{blue!6}}c}
\newcolumntype{d}{>{\columncolor{blue!6}}c}
\renewcommand{\arraystretch}{0.9} 
\setlength{\belowcaptionskip}{2pt} 
\vspace{-2em} 
\begin{table*}[t]
\footnotesize
  \centering
   \resizebox{\textwidth}{!}{%
    \begin{tabular}{cqbbbbb}
    \toprule
    \multirow{1}[0]{*}{\textbf{Domain}} & \multicolumn{1}{c}{\textbf{Pre-Update}} & \multicolumn{5}{c}
    {\textbf{Post-Update}}\\
    \cmidrule(lr){2-2}
    \cmidrule(lr){3-7}
      & \multicolumn{1}{c}{\textbf{$\mathcal{S}_{old}$}}  & \multicolumn{1}{c}{\textbf{SlideRegen}} & \multicolumn{1}{c}{\textbf{FullRegen}} & \multicolumn{1}{c}{\textbf{IncUpdate}} & \multicolumn{1}{c}{\textbf{HierMerge}} & \multicolumn{1}{c}{\textbf{\method}} \\
    \midrule
    \rowcolor[gray]{0.95} \multicolumn{7}{c}{\textit{Previous Window Prediction ($\varepsilon_{prev|\mathcal{S}_{old/new}}$) - \textbf{Previous Preservation}}} \\
    \midrule
        Recipe  & 0.57    &0.95\ \textcolor{gray}{(0.38$\uparrow$)}    &0.83\ \textcolor{gray}{(0.26$\uparrow$)}    &0.83\ \textcolor{gray}{(0.26$\uparrow$)}    &0.71\ \textcolor{gray}{(0.14$\uparrow$)}    &\underline{0.70}\ \textcolor{gray}{(0.13$\uparrow$)}    \\ 
        Book  & 0.78    &1.09\ \textcolor{gray}{(0.31$\uparrow$)}    &0.94\ \textcolor{gray}{(0.16$\uparrow$)}    &0.91\ \textcolor{gray}{(0.13$\uparrow$)}    &0.88\ \textcolor{gray}{(0.10$\uparrow$)}    &\underline{0.76}\ \textcolor{orange!60}{(0.02$\downarrow$)}    \\ 
        Clothing Shoes and jewelry  & 0.63    &1.13\ \textcolor{gray}{(0.50$\uparrow$)}    &0.96\ \textcolor{gray}{(0.33$\uparrow$)}    &0.94\ \textcolor{gray}{(0.31$\uparrow$)}    &\underline{0.77}\ \textcolor{gray}{(0.14$\uparrow$)}    &0.82\ \textcolor{gray}{(0.19$\uparrow$)}    \\ 
        Local Business  & 0.63    &1.10\ \textcolor{gray}{(0.47$\uparrow$)}    &0.95\ \textcolor{gray}{(0.32$\uparrow$)}    &0.91\ \textcolor{gray}{(0.28$\uparrow$)}    &0.74\ \textcolor{gray}{(0.11$\uparrow$)}    &\underline{0.73}\ \textcolor{gray}{(0.10$\uparrow$)}    \\ 
        Movies and TV  & 0.92    &1.17\ \textcolor{gray}{(0.25$\uparrow$)}    &1.03\ \textcolor{gray}{(0.11$\uparrow$)}    &1.00\ \textcolor{gray}{(0.08$\uparrow$)}    &0.98\ \textcolor{gray}{(0.06$\uparrow$)}    &\underline{0.85}\ \textcolor{orange!60}{(0.07$\downarrow$)}    \\ 
        MovieLens  & 0.76    &0.89\ \textcolor{gray}{(0.13$\uparrow$)}    &0.83\ \textcolor{gray}{(0.07$\uparrow$)}    &0.80\ \textcolor{gray}{(0.04$\uparrow$)}    &0.80\ \textcolor{gray}{(0.04$\uparrow$)}    &\underline{0.74}\ \textcolor{orange!60}{(0.02$\downarrow$)}    \\ 
        Arts Crafts and Sewing  & 0.49    & 0.81\ \textcolor{gray}{(0.32$\uparrow$)}    & 0.74\ \textcolor{gray}{(0.25$\uparrow$)}    & 0.68\ \textcolor{gray}{(0.19$\uparrow$)}    & 0.59\ \textcolor{gray}{(0.10$\uparrow$)}    &\underline{0.46}\ \textcolor{orange!60}{(0.03$\downarrow$)}    \\
        Automotive  & 0.55    &1.00\ \textcolor{gray}{(0.45$\uparrow$)}    &0.93\ \textcolor{gray}{(0.38$\uparrow$)}    &0.82\ \textcolor{gray}{(0.27$\uparrow$)}    &0.66\ \textcolor{gray}{(0.11$\uparrow$)}    &\underline{0.63}\ \textcolor{gray}{(0.08$\uparrow$)}    \\ 
        Sports and Outdoors  & 0.56  &0.99\ \textcolor{gray}{(0.43$\uparrow$)}    &0.87\ \textcolor{gray}{(0.31$\uparrow$)}    &0.85\ \textcolor{gray}{(0.29$\uparrow$)}    &0.67\ \textcolor{gray}{(0.11$\uparrow$)}    &\underline{0.66}\ \textcolor{gray}{(0.10$\uparrow$)}    \\ 
        Grocery and Gourmet Food  & 0.63    &1.13\ \textcolor{gray}{(0.50$\uparrow$)}    &1.00\ \textcolor{gray}{(0.37$\uparrow$)}    &0.95\ \textcolor{gray}{(0.32$\uparrow$)}    &\underline{0.71}\ \textcolor{gray}{(0.08$\uparrow$)}    &0.79\ \textcolor{gray}{(0.16$\uparrow$)}    \\

        \cdashlinelr{2-7}
        \textbf{Average} & \textbf{0.652} & \textbf{1.026} \textcolor{gray}{(0.374$\uparrow$)}  & \textbf{0.908} \textcolor{gray}{(0.256$\uparrow$)}  & \textbf{0.869} \textcolor{gray}{(0.217$\uparrow$)}  & \textbf{0.751} \textcolor{gray}{(0.099$\uparrow$)}  &\underline{\textbf{0.714}}\ \textcolor{gray}{(0.062$\uparrow$)}  \\

    \midrule
    \rowcolor[gray]{0.95} \multicolumn{7}{c}{\textit{Current Window Prediction ($\varepsilon_{curr|\mathcal{S}_{old/new}}$) - \textbf{Current Reflection}}} \\
    \midrule
        Recipe  &0.91    &0.78\ \textcolor{orange!60}{(0.13$\downarrow$)}    &0.84\ \textcolor{orange!60}{(0.07$\downarrow$)}    &\underline{0.41}\ \textcolor{orange!60}{(0.50$\downarrow$)}    &0.80\ \textcolor{orange!60}{(0.11$\downarrow$)}    & 0.44\ \textcolor{orange!60}{(0.47$\downarrow$)}    \\ 
        Book  &1.00    &0.92\ \textcolor{orange!60}{(0.08$\downarrow$)}    &0.97\ \textcolor{orange!60}{(0.03$\downarrow$)}    &0.41\ \textcolor{orange!60}{(0.59$\downarrow$)}    &0.91\ \textcolor{orange!60}{(0.09$\downarrow$)}    & \underline{0.35}\ \textcolor{orange!60}{(0.65$\downarrow$)}    \\ 
        Clothing Shoes and Jewelry  &1.00    &0.90\ \textcolor{orange!60}{(0.10$\downarrow$)}    &0.96\ \textcolor{orange!60}{(0.04$\downarrow$)}    &\underline{0.48}\ \textcolor{orange!60}{(0.52$\downarrow$)}    &0.90\ \textcolor{orange!60}{(0.10$\downarrow$)}    & 0.51\ \textcolor{orange!60}{(0.49$\downarrow$)}    \\ 
        Local Business  &1.04    &0.9\ \textcolor{orange!60}{(0.14$\downarrow$)}    &0.99\ \textcolor{orange!60}{(0.05$\downarrow$)}    &\underline{0.29}\ \textcolor{orange!60}{(0.75$\downarrow$)}    &0.93\ \textcolor{orange!60}{(0.11$\downarrow$)}    & 0.36\ \textcolor{orange!60}{(0.68$\downarrow$)}    \\ 
        Movies and TV  &1.12    &1.00\ \textcolor{orange!60}{(0.12$\downarrow$)}    &1.07\ \textcolor{orange!60}{(0.05$\downarrow$)}    &0.47\ \textcolor{orange!60}{(0.65$\downarrow$)}    &1.02\ \textcolor{orange!60}{(0.10$\downarrow$)}    & \underline{0.45}\ \textcolor{orange!60}{(0.67$\downarrow$)}    \\ 
        MovieLens  &0.87    &0.78\ \textcolor{orange!60}{(0.09$\downarrow$)}    &0.82\ \textcolor{orange!60}{(0.05$\downarrow$)}    &\underline{0.30}\ \textcolor{orange!60}{(0.57$\downarrow$)}    &0.80\ \textcolor{orange!60}{(0.07$\downarrow$)}    & 0.43\ \textcolor{orange!60}{(0.44$\downarrow$)}    \\ 
        Arts Crafts and Sewing  &0.76    &0.76\ \textcolor{gray}{(0.00$\uparrow$)}    &0.77\ \textcolor{gray}{(0.01$\uparrow$)}    &0.39\ \textcolor{orange!60}{(0.37$\downarrow$)}    &0.72\ \textcolor{orange!60}{(0.04$\downarrow$)}    & \underline{0.26}\ \textcolor{orange!60}{(0.50$\downarrow$)}    \\ 
        Automotive  &0.84    &0.81\ \textcolor{orange!60}{(0.03$\downarrow$)}    &0.88\ \textcolor{gray}{(0.04$\uparrow$)}    &0.38\ \textcolor{orange!60}{(0.46$\downarrow$)}    &0.81\ \textcolor{orange!60}{(0.03$\downarrow$)}    & \underline{0.27}\ \textcolor{orange!60}{(0.57$\downarrow$)}    \\ 
        Sports and Outdoors  &0.91  &0.79\ \textcolor{orange!60}{(0.12$\downarrow$)}    &0.84\ \textcolor{orange!60}{(0.07$\downarrow$)}    &0.37\ \textcolor{orange!60}{(0.54$\downarrow$)}    &0.82\ \textcolor{orange!60}{(0.09$\downarrow$)}    & \underline{0.36}\ \textcolor{orange!60}{(0.55$\downarrow$)}    \\ 
        Grocery and Gourmet Food  &1.19    &0.93\ \textcolor{orange!60}{(0.26$\downarrow$)}    &1.08\ \textcolor{orange!60}{(0.11$\downarrow$)}    &\underline{0.46}\ \textcolor{orange!60}{(0.73$\downarrow$)}    &1.05\ \textcolor{orange!60}{(0.14$\downarrow$)}    & 0.49\ \textcolor{orange!60}{(0.70$\downarrow$)}    \\

        \cdashlinelr{2-7}
        \textbf{Average} &\textbf{0.964} &\textbf{0.857}\ \textcolor{orange!60}{(0.107$\downarrow$)}  &\textbf{0.922}\ \textcolor{orange!60}{(0.042$\downarrow$)}  &\textbf{0.396}\ \textcolor{orange!60}{(0.568$\downarrow$)}  &\textbf{0.876}\ \textcolor{orange!60}{(0.088$\downarrow$)}  & \underline{\textbf{0.392}}\ \textcolor{orange!60}{(0.572$\downarrow$)}  \\ 
    
    \midrule
    \rowcolor[gray]{0.95} \multicolumn{7}{c}{\textit{Future Window Prediction ($\varepsilon_{fut|\mathcal{S}_{old/new}}$) - \textbf{Future Advancement}}} \\
    \midrule
        Recipe  & 0.91  & 0.92\ \textcolor{gray}{(0.01$\uparrow$)}   & 0.92\ \textcolor{gray}{(0.01$\uparrow$)}   & 0.91\ \textcolor{gray}{(0.00$\uparrow$)}   & 0.94\ \textcolor{gray}{(0.03$\uparrow$)}   & \underline{0.72}\ \textcolor{orange!60}{(0.19$\downarrow$)}   \\ 
        Book  & 1.01  & 1.06\ \textcolor{gray}{(0.05$\uparrow$)}   & 1.03\ \textcolor{gray}{(0.02$\uparrow$)}   & 0.96\ \textcolor{orange!60}{(0.05$\downarrow$)}   & 1.03\ \textcolor{gray}{(0.02$\uparrow$)}   & \underline{0.79}\ \textcolor{orange!60}{(0.22$\downarrow$)}   \\ 
        Clothing Shoes and Jewelry  & 1.03  & 1.09\ \textcolor{gray}{(0.06$\uparrow$)}   & 1.03\ \textcolor{gray}{(0.00$\uparrow$)}   & 1.00\ \textcolor{orange!60}{(0.03$\downarrow$)}   & 1.04\ \textcolor{gray}{(0.01$\uparrow$)}   & \underline{0.88}\ \textcolor{orange!60}{(0.15$\downarrow$)}   \\ 
        Local Business  & 1.04  & 1.06\ \textcolor{gray}{(0.02$\uparrow$)}   & 1.04\ \textcolor{gray}{(0.00$\uparrow$)}   & 0.97\ \textcolor{orange!60}{(0.07$\downarrow$)}   & 1.04\ \textcolor{gray}{(0.00$\uparrow$)}   & \underline{0.80}\ \textcolor{orange!60}{(0.24$\downarrow$)}   \\ 
        Movies and TV  & 1.18  & 1.14\ \textcolor{orange!60}{(0.04$\downarrow$)}   & 1.12\ \textcolor{orange!60}{(0.06$\downarrow$)}   & 1.06\ \textcolor{orange!60}{(0.12$\downarrow$)}   & 1.12\ \textcolor{orange!60}{(0.06$\downarrow$)}   & \underline{0.98}\ \textcolor{orange!60}{(0.20$\downarrow$)}   \\ 
        MovieLens  & 0.85  & 0.84\ \textcolor{orange!60}{(0.01$\downarrow$)}   & 0.83\ \textcolor{orange!60}{(0.02$\downarrow$)}   & 0.76\ \textcolor{orange!60}{(0.09$\downarrow$)}   & 0.82\ \textcolor{orange!60}{(0.03$\downarrow$)}   & \underline{0.73}\ \textcolor{orange!60}{(0.12$\downarrow$)}   \\ 
        Arts Crafts and Sewing  & 0.75  & 0.81\ \textcolor{gray}{(0.06$\uparrow$)}   & 0.77\ \textcolor{gray}{(0.02$\uparrow$)}   & 0.71\ \textcolor{orange!60}{(0.04$\downarrow$)}   & 0.75\ \textcolor{gray}{(0.00$\uparrow$)}   & \underline{0.43}\ \textcolor{orange!60}{(0.32$\downarrow$)}   \\ 
        Automotive  & 0.86  & 0.96\ \textcolor{gray}{(0.10$\uparrow$)}   & 0.92\ \textcolor{gray}{(0.06$\uparrow$)}   & 0.88\ \textcolor{gray}{(0.02$\uparrow$)}   & 0.90\ \textcolor{gray}{(0.04$\uparrow$)}   & \underline{0.61}\ \textcolor{orange!60}{(0.25$\downarrow$)}   \\ 
        Sports and Outdoors  & 0.97  & 0.94\ \textcolor{orange!60}{(0.03$\downarrow$)}   & 0.93\ \textcolor{orange!60}{(0.04$\downarrow$)}   & 0.89\ \textcolor{orange!60}{(0.08$\downarrow$)}   & 0.90\ \textcolor{orange!60}{(0.07$\downarrow$)}   & \underline{0.80}\ \textcolor{orange!60}{(0.17$\downarrow$)}   \\ 
        Grocery and Gourmet Food  & 1.25  & 1.14\ \textcolor{orange!60}{(0.11$\downarrow$)}   & 1.20\ \textcolor{orange!60}{(0.05$\downarrow$)}   & 1.10\ \textcolor{orange!60}{(0.15$\downarrow$)}   & 1.19\ \textcolor{orange!60}{(0.06$\downarrow$)}   & \underline{0.89}\ \textcolor{orange!60}{(0.36$\downarrow$)}   \\ 
        
        \cdashlinelr{2-7}
        \textbf{Average} & \textbf{0.985}  & \textbf{0.996}\ \textcolor{gray}{(0.011$\uparrow$)}   & \textbf{0.979}\ \textcolor{orange!60}{(0.006$\downarrow$)}   & \textbf{0.924}\ \textcolor{orange!60}{(0.061$\downarrow$)}   & \textbf{0.973}\ \textcolor{orange!60}{(0.012$\downarrow$)}   & \underline{\textbf{0.763}}\ \textcolor{orange!60}{(0.222$\downarrow$)}   \\ 
    \bottomrule
    \end{tabular}%
     }
     \caption{MAE results of previous, current, and future window prediction tasks using personas Pre- and Post- the first update with different methods. This table illustrates how well each method achieves the three high-level goals: Previous Preservation, Current Reflection, and Future Advancement. It presents the changes in MAE ($\left| \varepsilon_{t | \mathcal{S}_{old}}-\varepsilon_{t | \mathcal{S}_{new}} \right|$) relative to the old persona, with upward arrows (\textcolor{gray}{$\uparrow$}) indicating error increases and downward arrows (\textcolor{orange!60}{$\downarrow$}) indicating error reductions. Average results are highlighted in \textbf{bold}, and the best results are \underline{underlined}}
     \captionsetup{skip=2pt} 
  \label{tab:goals}
\end{table*}
\FloatBarrier 
\vspace{-2em} 

\paragraph{\method excels in identifying high-quality directions.}
Building on previous insights,
we further evaluate \method’s ability to identify high-quality refinement directions by analyzing its performance across three goals (Table~\ref{tab:goals}). 
\method demonstrates outstanding performance: minimizing previous forgetting with the smallest average MAE increment of 0.062 (\textit{Previous Preservation}); reducing current errors by 0.572 on average (\textit{Current Reflection}); and improving future predictions with an average 
reduction of 0.222 (\textit{Future Advancement}). Notably, \method surpasses all baselines across domains for \textit{Future Advancement}, demonstrating its capacity step-wise optimization. These results highlight \method’s ability to balance three goals for better direction search and continual optimization.

\begin{figure}[t]
    \centering
    \includegraphics[width=\linewidth]{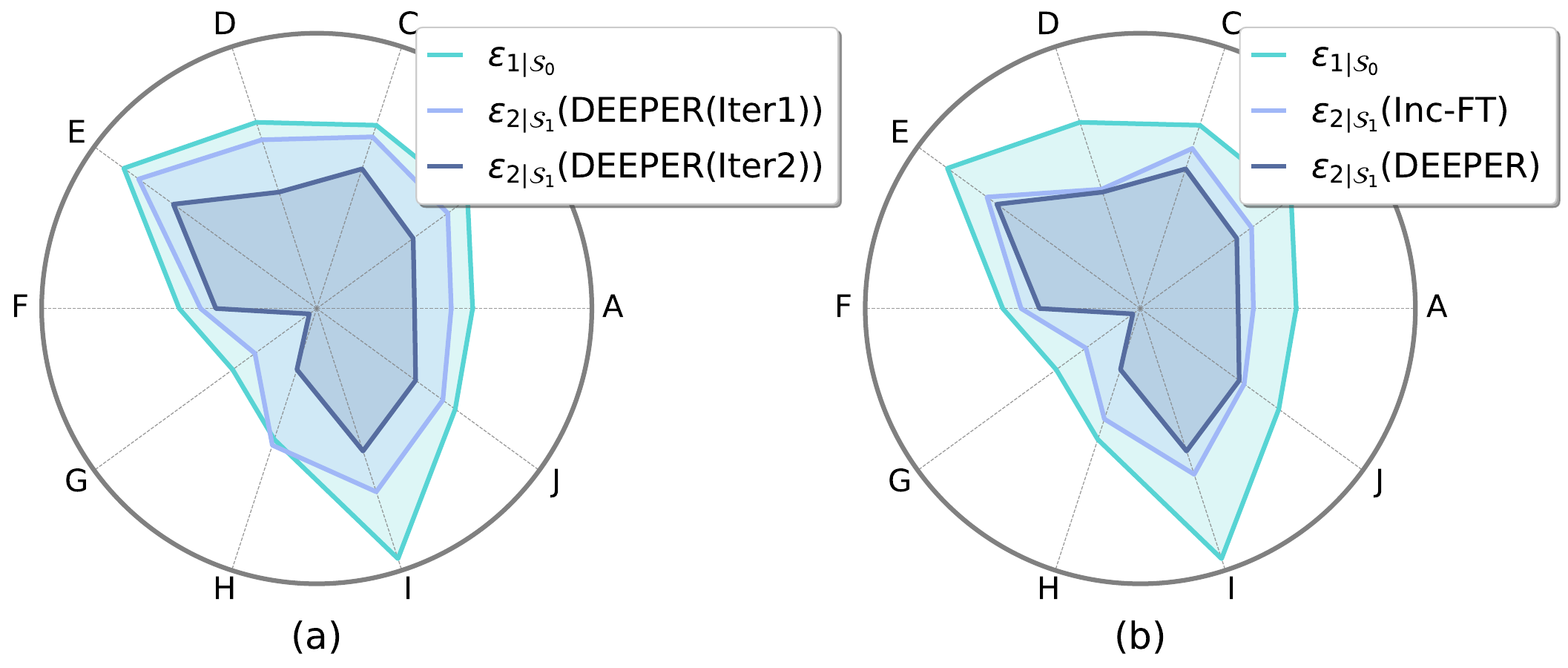}
    \caption{(a) Refinement performance across two offline RL iterations of \method. (b) Comparison of \method and fine-tuned \textit{IncUpdate} (Inc-FT). }
    \label{fig:analysis_radar_part2}
\end{figure}

\paragraph{Iterative training enhances persona refinement.}
Guided by stage-specific objectives, \method’s two-stage iterative RL framework incrementally enhances refinement capabilities by leveraging progressively higher-quality self-sampled data and expanded preference margins. Results (Figure~\ref{fig:analysis_radar_part2}(a)) show accelerated improvements in the second iteration, highlighting effects of iterative training.

\paragraph{Prediction discrepancy facilitates direction search.}  
We finally analyze paradigm's role in direction search by employing \method’s training framework into \textit{IncUpdate} (the best-performing baseline). Figure~\ref{fig:analysis_radar_part2}(b) show that while direction search training improves \textit{IncUpdate}’s performance, it still falls short of \method. This underscores prediction discrepancy’s role in enabling context-specific search and more precise refinement.

\subsection{Persona Probing}
We further conduct an preliminary analysis of refined personas, termed \textit{persona probing}, to explore additional insights and applications of \method.

\renewcommand{\arraystretch}{0.9} 

\begin{table}[htbp]
\centering
\footnotesize

\begin{tabular}{lccccc}
\toprule
\textbf{Update Method} & \textbf{$\mathcal{S}_0$} & \textbf{$\mathcal{S}_1$} & \textbf{$\mathcal{S}_2$} & \textbf{$\mathcal{S}_3$} & \textbf{$\mathcal{S}_4$}\\
\midrule
\method & 245.0 & 316.8 & 353.5 & 393.2 & 429.4\\
IncUpdate & 245.0 & 390.1 & 459.3 & 500.4 & 526.4\\
HierMerge & 245.0 & 325.3 & 393.5 & 462.2 & 509.1\\
\bottomrule
\end{tabular}
\caption{Average persona token count across rounds.}
\label{tab:persona_tokens} 
\end{table}

\begin{figure}[h]
    \centering
    \includegraphics[width=0.95\linewidth]{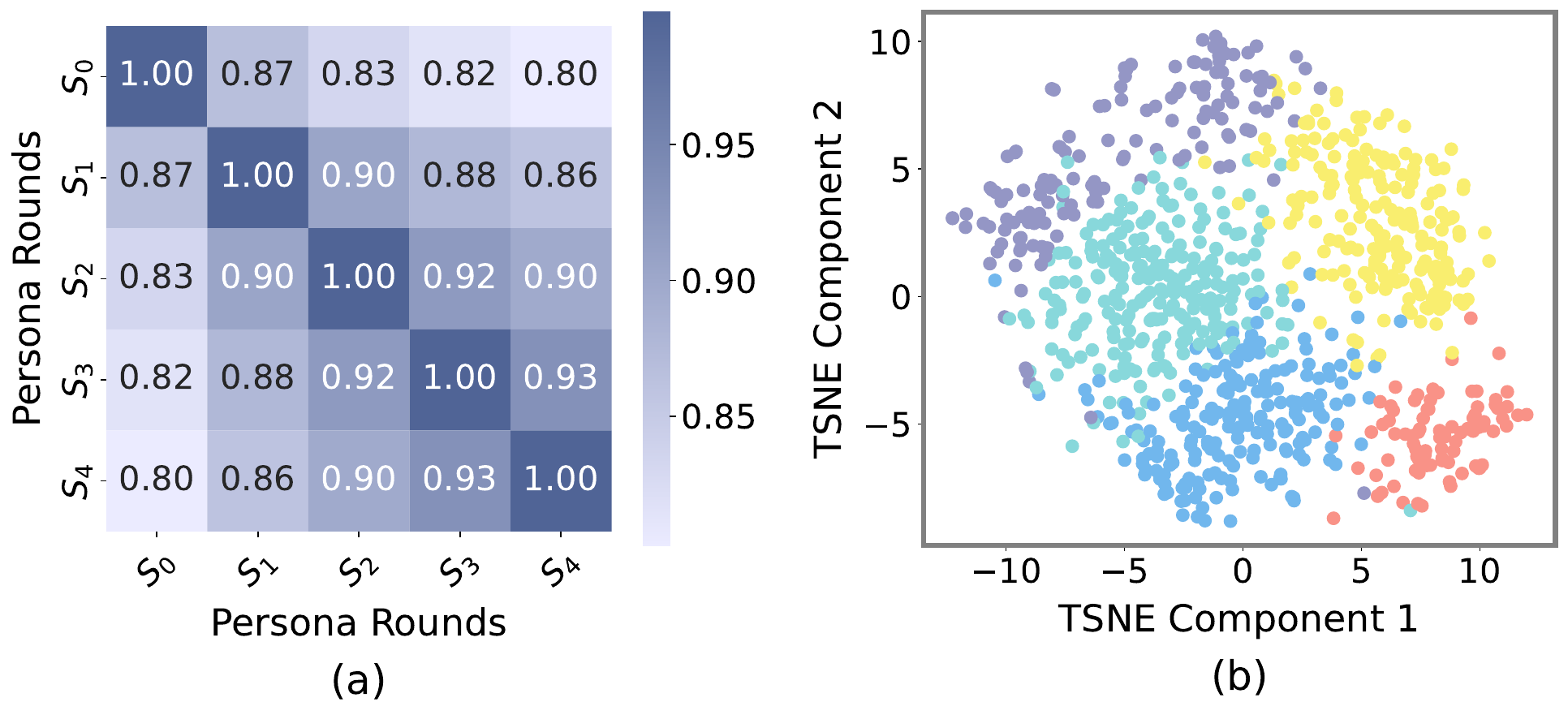}
    \caption{(a) Cosine similarity among personas across rounds; (b) User clusters based on final personas(Book).}
    \label{fig:persona_probing}
\end{figure}

\noindent
\textbf{Dynamic persona evolution across rounds.}
We first analyze persona dynamics during refinement process. Table~\ref{tab:persona_tokens} highlights \method’s controlled length growth, balancing representation efficiency and informativeness. Figure~\ref{fig:persona_probing}(a) reveals diminishing persona changes over time, with substantial shifts in early updates ($\mathcal{S}_0 \to \mathcal{S}_1$) and increasing stability in later rounds ($\mathcal{S}_1 \to \mathcal{S}_4$), indicating convergence and improved contextual alignment.

\renewcommand{\arraystretch}{0.9} 

\begin{table}[h]
\centering
\footnotesize
\begin{tabular}{lc}
\toprule
Profiling Dimensions(Book) & User Count \\
\midrule
Story \& Plot & 871\\
Emotion \& Experience & 878\\
Genre \& Theme & 878\\
Social \& Cultural Context & 680\\
User Behavior Traits & 862\\
Author \& Character & 701\\
Personality \& Values & 867\\
Relationship \& Connection & 716\\
\bottomrule
\end{tabular}
\caption{Key profiling dimensions in Book domain.}
\label{tab:book_dimensions} 
\end{table}

\paragraph{Insights from final optimized personas.} 
Refined personas from \method also enable in-depth, domain-specific exploration. 
In Book domain, we uncover \textbf{group-level preferences} by clustering final persona embeddings, identifying 5 user groups characterized by unique high-frequency adjectives (e.g., “romantic” and “practical”) (Figure~\ref{fig:persona_probing}(b)). 
We also extract \textbf{domain-specific patterns} by organizing high-frequency terms into 8 dimensions using GPT-4o~\cite{gpt4} (Table~\ref{tab:book_dimensions}), highlighting critical factors for modeling Book domain users. These attempts show \method’s potential to support strategic user insights exploration.

\section{Related Work}
\label{sec:related}
\paragraph{Persona Modeling} 
Persona modeling in personalized applications captures user preferences and behaviors from behavioral data or dialogue history, with advancements driven by LLMs \cite{li2021survey, tan2023user, tseng2024two}. Most studies focus on one-time persona generation from static user behavior or profile data \cite{ji2023chatgpt, wang2023zero, zhou2024language, wu2024rlpf, wang2024incharacter, xu2024character, lyu2023llm, salemi2023lamp}. To address real-world challenges, dynamic persona modeling using streaming user data has emerged \cite{lian2022incremental, wang2020incremental, yin2023heterogeneous, qin2024enhancing}. Departing from regeneration- and extension-based approaches, our method refines personas by integrating user behaviors and model predictions for more accurate and effective updates.

\paragraph{LLM for Recommendation and Behavior Prediction}
LLMs are increasingly applied in personalized systems like recommendation engines \cite{wang2023recmind,wu_surveylargelanguagemodels_2024,zhang2023recommendation}. Some studies integrate LLMs into traditional frameworks to enhance user modeling and contextual understanding \cite{liu2024once,zhang2024generative,li2023ctrl, li2023exploring}, while others employ LLMs directly for generating recommendations or predicting future behaviors, leveraging their persona modeling capabilities for greater adaptability and precision \cite{liu2023chatgpt, lyu2023llm, gao2023chat,dai2023uncovering}. This work leverages LLMs for dynamic persona modeling and behavior prediction to capture users’ evolving preferences.

\section{Conclusion}
\label{sec:conclusion}
In this paper, we introduce \method, an effective approach to dynamic persona modeling that leverages iterative offline RL and discrepancy-based refinement to continuously enhance persona quality and predictive accuracy. Comprehensive experiments demonstrate \method’s effectiveness across diverse domains in dynamic user modeling. We hope \method marks a significant advancement in personalized applications.

\section*{Limitations}
\label{sec:limitation}
First, this study focuses on dynamic persona modeling using discrete, quantifiable user behaviors, such as ratings, as \method relies on prediction discrepancies for updates and reward computation. Other data forms, such as natural language interactions, are beyond its scope. Second, due to data availability constraints, we validate \method using user rating prediction tasks, which are widely applicable and provide ample real-world sequential behavior data across domains. Nevertheless, the \method framework is adaptable to broader user interaction scenarios. Finally, the insights derived from the book domain are specific to the dataset and model used, and their generalizability to other datasets and models remains uncertain.

\section*{Ethics Statement}
\label{sec:Ethics}
\paragraph{Risks} First, the datasets used in this work are publicly available and anonymized. However, we acknowledge that user behavior data, even in aggregate form, may raise privacy concerns if not handled properly. Second, our model relies on datasets that may not fully represent all user groups or domains, leading to potential biases in persona refinement and prediction. The proposed method could potentially be misused for excessive user behavior tracking or manipulative personalization. Developers and practitioners should ensure ethical use in line with user privacy regulations.

\section*{Acknowledgement}
\label{sec:Acknowledge}
Aili Chen, Yanghua Xiao et al. gratefully acknowledge the research funding provided by Alibaba Group. The authors also extend their sincere appreciation to Xintao Wang from Fudan University for his insightful feedback and valuable contributions throughout the research process.

\bibliography{custom}

\clearpage
\appendix

\section{Dynamic Persona Modeling Details}
\subsection{Persona Initialization}
In this study, we employ the frozen LLM, GPT-4o-mini, to initialize user personas based on their first 10 ratings ($\mathcal{W}_0$) during the initial stage of dynamic persona modeling. The prompt used for persona initialization is presented in Table~\ref{persona_initialize}.

\begin{tcolorbox}
{\fontfamily{cmtt}\selectfont\small
TASK: Infer the user's persona based \\
on their ratings of {item\_type} items.\\
Instructions: \\
Below is a list of \\
\{item\_type\}s that the user has rated. \\
Each rating ranges from 1 to 5:\\
\{user\_ratings\}\\

Based on these, generate a user persona without mentioning item names or rating scores. 
\\
User Persona(at least  **200 words**):}
\end{tcolorbox}
\noindent\begin{minipage}{0.48\textwidth}
\captionof{table}{Persona Initialization prompt template.}\label{persona_initialize}
\end{minipage}

\subsection{Behavior Observation and Prediction}
For all user behavior prediction task, we use GPT-4o-mini to role-play the given input persona and predict user ratings on the given item list. Table~\ref{persona_prediction} shows the prompt template for the prediction task.

\begin{tcolorbox}
{\fontfamily{cmtt}\selectfont\small
TASK: Role-play the given persona and predict what score (out of 5) you would give to the following \{item\_type\} list.\\
Instructions: Based on the persona \{persona\}, predict ratings for each item in the list below.\\
\{items\}\\
\#\# Output format:\\
```json\\
\texttt{[}\\
    \{\{"item\_name":..., "predict\_rating":...\}\},\\
    \{\{"item\_name":..., "predict\_rating":...\}\},\\
    ...\\
\texttt{]}\\
```}
\end{tcolorbox}
\noindent\begin{minipage}{0.48\textwidth}
\captionof{table}{Behavior Prediction prompt template.}\label{persona_prediction}
\end{minipage}

\subsection{Persona Update}
In this work, the formulation of persona update stage varies by corresponding paradigms. For all baselines, we use GPT-4o-mini as backbone, while for \method, we use the fine-tuned model via the iterative offline RL training framework based on Llama3.1-8b-Instruct. For each persona update method, we design corresponding update prompt template as follows.

\textbf{\method }
    The \method approach based on a refinement-based paradigm with predicted and actual user ratings. Table~\ref{persona_refinement_method_deeper} shows prompt template for persona update with \method.
    
    \begin{tcolorbox} 
        {\fontfamily{cmtt}\selectfont\small 
        TASK: \\
        Refine the old user persona based on differences 
        between predicted and actual ratings of \{item\_type\} 
        items.\ Instructions: \\
        Below is the existing persona 
        inferred from past behavior:\\
        \{old\_persona\} \\
        Below is the comparison of predicted ratings (based on the old persona) 
        versus actual ratings:\\
        \{predict\_and\_actual\_user\_ratings\}\\
        Reflect on these differences and generate a refined 
        user persona without mentioning item names or rating scores.
        Refined User Persona:

        }
        \end{tcolorbox}
        \noindent\begin{minipage}{0.48\textwidth}
        \captionof{table}{\method Persona Refinement prompt template.}\label{persona_refinement_method_deeper}
        \end{minipage}

\textbf{FullRegen}
In the FullRegen, we fully regenerate the user's persona whenever new ratings are provided. This method does not consider the prior persona and instead creates a fresh representation based all observed ratings. Table~\ref{persona_update_fullregen} shows the prompt template for persona update with FullRegen.

\begin{tcolorbox}
    {\fontfamily{cmtt}\selectfont\small
    TASK: Infer the user's persona based \\
    on their ratings of {item\_type} items.\\
    Instructions: \\
    Below is a list of \\
    \{item\_type\}s that the user has rated. \\
    Each rating ranges from 1 to 5:\\
    \{Full\_user\_ratings\}\\
    
    Based on these, generate a user persona without mentioning item names or rating scores. 
\\
    User Persona:
}
    \end{tcolorbox}
\noindent\begin{minipage}{0.48\textwidth}
\captionof{table}{FullRegen Persona Update prompt template.}
\label{persona_update_fullregen}
\end{minipage}

\textbf{SlideRegen} 
In the SlideRegen method, we regenerate personas based on their recent ratings of 
\{item\_type\} items(latest window). 
Table~\ref{persona_update_slideregen} shows the prompt template for persona update with SlideRegen.

\begin{tcolorbox}
    {\fontfamily{cmtt}\selectfont\small
    TASK: Infer the user's persona based \\
    on their ratings of {item\_type} items.\\
    Instructions: \\
    Below is a list of \\
    \{item\_type\}s that the user has rated. \\
    Each rating ranges from 1 to 5:\\
    \{Slide\_user\_ratings\}\\\\
Based on these, generate a user persona without mentioning item names or rating scores. 
\\
    User Persona:
   }
\end{tcolorbox}
\noindent\begin{minipage}{0.48\textwidth}
\captionof{table}{SlideRegen Persona Update prompt template.}\label{persona_update_slideregen}
\end{minipage}

\textbf{IncUpdate}
In the IncUpdate, the user's persona is dynamically updated by integrating new ratings with their existing persona. Table~\ref{persona_update_incupdate} shows prompt template for persona update with IncUpdate.

\begin{tcolorbox} {\fontfamily{cmtt}\selectfont\small 
    TASK: 
    Integrate the user's most recent ratings of \{item\_type\} \\
    items into their existing persona to generate an updated persona. \\
    Instructions: \\
    Below is the existing persona based on prior behaviors: \\
    \{old\_persona\} \\
    Below is a list of recent \{item\_type\}s that the user has rated.\\
    \ Each rating ranges from 1 to 5:\\
    \{user\_ratings\}\\
    \ Based on these, integrate 
    the new features from the recent ratings into the existing persona.\\
    \ Updated Persona:
   }
    \end{tcolorbox}
    \noindent\begin{minipage}{0.48\textwidth}
    \captionof{table}{IncUpdate Persona Update prompt template.}\label{persona_update_incupdate}
    \end{minipage}

\textbf{HierMerge}  
    The HierMerge method combines both long-term personas and short-term personas hierarchically. 
    Table~\ref{persona_update_hiermerge} shows prompt template for persona update with HierMerge.
    \begin{tcolorbox}
    {\fontfamily{cmtt}\selectfont\small
    \# Prompt 1:\\
    TASK: Infer the user's persona based on \\
    their ratings of \{item\_type\} items.\\
    Instructions: \\
    Below is a list of \\
    \{item\_type\}s that the user has rated.\\
    Each rating ranges from 1 to 5:\\
    \{user\_ratings\}\\
    Based on these, generate a user\\
     persona without mentioning item names or rating scores.\\
    User Persona:\\

    \# Prompt 2:
    TASK: 
    Update the long-term persona by merging 
    it with the newly generated short-term 
    persona.
    Instructions: \\
    Below is 
    the existing long-term persona based 
    on prior behaviors:\{long\_term\_persona\} \\
    Below is the newly generated short-term persona 
    based on recent behaviors:\{short\_term\_persona\} \\ 
    Merge the short-term persona into the long-term persona 
    to capture both historical stability and recent dynamics.\\ 
    The updated persona should reflect both long-term \\
    preferences and recent changes without losing consistency.\\ 
    Updated Long-Term Persona: \\
   
    }
    \end{tcolorbox}
    \noindent\begin{minipage}{0.48\textwidth}
    \captionof{table}{HierMerge Persona Update prompt template.}\label{persona_update_hiermerge}
    \end{minipage}

\begin{table*}[t]
    \footnotesize
    \begin{tabular}{cccccc}
    \toprule
    \midrule
    \textbf{Dataset} & \textbf{Abbreviation} & \textbf{Usage} & \multicolumn{1}{c}{\begin{tabular}[c]{@{}c@{}}\textbf{\# Users}\\ \textbf{in Train}\\ \end{tabular}} & \multicolumn{1}{c}{\begin{tabular}[c]{@{}c@{}}\textbf{\# Users}\\ \textbf{in Eval}\end{tabular}} & \multicolumn{1}{c}{\begin{tabular}[c]{@{}c@{}}\textbf{\# Domain}\\ \textbf{Label}\end{tabular}} \\ \midrule
    \begin{tabular}[c]{@{}c@{}}Food.com Recipes \\ - and Interactions\end{tabular} & Recipe & Train/Eval & 1000 & 356 & A \\  \midrule
    \begin{tabular}[c]{@{}c@{}}Amazon Review Data (2018) \\ - Books\end{tabular} & Book & Train/Eval & 3000 & 897 & B \\\midrule  
    \begin{tabular}[c]{@{}c@{}}Amazon Review Data (2018) \\ - Clothing Shoes and Jewelry\end{tabular} & Clothing Shoes and Jewelry & Train/Eval & 300 & 243 & C \\  \midrule
    \begin{tabular}[c]{@{}c@{}}Google Local Data (2021) \\ - New York\end{tabular} & Local Business & Train/Eval & 2500 & 826 & D \\ \midrule
    \begin{tabular}[c]{@{}c@{}}Amazon Review Data (2018)\\ - Movies and TV\end{tabular} & Movies and TV & Train/Eval & 1000 & 837 & E \\ \midrule 
    \begin{tabular}[c]{@{}c@{}}MovieLens 20M Dataset\end{tabular} & MovieLens & Train/Eval & 3000 & 1000 & F \\ \midrule 
    \begin{tabular}[c]{@{}c@{}}Amazon Review Data (2018) \\ - Art Crafts and Sewing\end{tabular} & Art Crafts and Sewing & Eval & - & 86 & G \\  \midrule
    \begin{tabular}[c]{@{}c@{}}Amazon Review Data (2018) \\ - Automative\end{tabular} & Automative & Eval & - & 143 & H \\  \midrule
    \begin{tabular}[c]{@{}c@{}}Amazon Review Data (2018) \\ - Sports and Outdoors\end{tabular} & Sports and Outdoors & Eval & - & 236 & I \\  \midrule
    \begin{tabular}[c]{@{}c@{}}Amazon Review Data (2018) \\ - Grocery and Gourmet Food\end{tabular} & Grocery and Gourmet Food & Eval & - & 185 & J \\ 
    \bottomrule
    \end{tabular}
    \caption{Details of Datasets Used in Experiments.}
    \label{tab:data1set}
    \vskip -0.1in
\end{table*}

\subsection{User Details}

We utilize four publicly available and well-known datasets, selecting a total of 10 domains. From six domains, we randomly sampled a total of 14,959 users with at least 50 ratings. Among these, 10,800 users are used for constructing the training data, and 4,159 users are used for constructing the testing data. Additionally, to evaluate the generalization ability of the methods, we sampled 650 users with at least 50 ratings from four unseen domains to construct an additional test set. Each user's 50 rating behaviors are sorted by timestamp and divided into five sequences of length 10, simulating multi-round online user interactions. The detailed user sampling statistics are in Table ~\ref{tab:data1set}:

\subsection{Training Data Construction}
In \textbf{Iteration 1}, a total of 10,800 
context data points are constructed, 
each corresponding to the first persona refinement step for each user. 
For each context, 15 candidate personas are randomly sampled using the 
Llama3.1-8b-Instruct model, with inference parameters set as follows: 
\texttt{temperature=1} (to ensure diversity among the candidates), 
\texttt{top\_p=0.4} (to control the cumulative probability of tokens), 
and \texttt{repetition\_penalty=1.1} (to prevent repetition in the generated output). 
The boundaries for positive and negative reward sets are set to 0.5 and 0, 
with a margin of 0.5. In total, 34,782 DPO preference pairs are constructed, 
with 10\% randomly selected for the validation set. This data is used to train 
\textbf{Model 1}.

In \textbf{Iteration 2}, \textbf{Model 1} is first used to generate outputs for the 10,800 context data points from Iteration 1, completing the first persona update for each user. These results, in turn, are used to construct a second set of 10,800 context data points for the second persona refinement. These are then combined with the 10,800 context data points constructed in the first iteration, resulting in a total of 21,600 context data points for sampling in the second iteration. For each context, 15 candidate personas are randomly sampled using \textbf{Model 1}, with the same inference parameters as in Iteration 1. The boundaries for positive and negative reward sets are set to 0.5 and 0, with a margin of 0.8. A total of 28,612 new DPO preference pairs are generated. Additionally, 5,000 preference pairs with a margin greater than 0.8 are randomly selected from Iteration 1 to be included in the training set, mitigating the issue of catastrophic forgetting. This results in a total of 33,612 DPO preference pairs, with 10\% randomly selected for the validation set, used to train \textbf{Model 2}.

\begin{table*}[ht]
   \footnotesize
    \centering
    \caption{Hyperparameter Details for Training}
    \label{tab:hyperparameters}
    \begin{tabular}{ccc}
        \toprule
        \midrule
    \textbf{Parameter}               & \textbf{Value}                                  & \textbf{Description}                                  \\
    \\\midrule
    Model Name or Path               & \texttt{Llama-3.1-8B-Instruct}                  & Path to the model                                     \\
    Finetuning Type                  & \texttt{lora}                                   & Type of finetuning                                   \\ 
    Training Stage                   & \texttt{dpo}                                    & Current training stage                               \\ 
    LoRA Target                      & \texttt{all}                                    & LoRA target layers                                  \\  
    LoRA Rank                        & 16                                              & LoRA rank                                           \\  
    LoRA Alpha                       & 32                                              & LoRA alpha                                          \\  
    LoRA Dropout                     & 0.2                                             & LoRA dropout rate                                   \\  
    Preference Beta                  & 0.2                                             & Preference loss beta                                \\  
    Preference Loss Type             & \texttt{sigmoid}                                & Type of preference loss                             \\  
    Preference Finetune Rate         & 0.1                                             & Preference finetuning rate                         \\ 
    Maximum Sequence Length          & 2048                                            & Maximum input sequence length                      \\  
    Training Batch Size              & 4                                               & Batch size per device during training             \\  
    Gradient Accumulation Steps      & 8                                               & Steps for gradient accumulation                   \\  
    Learning Rate                    & 5.0e-06                                         & Learning rate                                     \\  
    Number of Epochs                 & 4.0                                             & Total number of training epochs                   \\  
    Learning Rate Scheduler          & \texttt{cosine}                                 & Learning rate scheduling strategy                 \\  
    Warmup Steps                     & 250                                             & Warmup steps before full learning rate            \\  
    Maximum Gradient Norm            & 1.0                                             & Maximum norm for gradient clipping                \\  
    BF16 Precision                   & \texttt{true}                                   & Use BF16 precision                                 \\ 
    Optimizer                        & \texttt{adamw\_torch}                           & Type of optimizer                                 \\  
    Validation Size                  & 0.1                                             & Fraction of data used for validation              \\  
    Evaluation Batch Size            & 4                                               & Batch size per device during evaluation           \\  
    Evaluation Strategy              & \texttt{steps}                                  & Evaluation scheduling strategy                    \\  
    Evaluation Steps                 & 100                                             & Steps between evaluations                         \\  
    \midrule
    \bottomrule    
\end{tabular}
    \end{table*}

\newpage
\section{Training Details}

\subsection{Hyperparameter Details}

Our training pipeline is implemented based on LlamaFactory ~\cite{zheng2024llamafactory} and the hyperparameters used in the training process are summarized in Table~\ref{tab:hyperparameters}.

\subsection{Loss Function Details}
In this section, we provide the detailed formulations of the training loss functions combined with DPO loss and SFT loss.

\paragraph{DPO Loss}
The DPO loss optimizes the model by leveraging user preference signals to align persona refinements with higher rewards. 

\begin{align}
  \small
  \mathcal{L}_{\text{DPO}}(\pi_\theta; \pi_{\text{ref}}) = 
  &-\mathbb{E}_{(x, y_w, y_l) \sim \mathcal{D}} \notag \\
  &\left[ 
  \log \sigma \left( 
  \beta \log \frac{\pi_\theta(y_w \mid x)}{\pi_{\text{ref}}(y_w \mid x)} 
  \right. \right. \notag \\
  &\left. \left. - 
  \beta \log \frac{\pi_\theta(y_l \mid x)}{\pi_{\text{ref}}(y_l \mid x)} 
  \right) 
  \right]. \tag{11}
  \end{align}

The policy distribution of the model being trained, parameterized by \(\theta\). It represents the probability of generating specific outputs conditioned on the input \(x\).
The reference model's policy distribution, used as a baseline for comparison. It is typically a pretrained model or a checkpoint used to stabilize training.
A tuple sampled from the dataset \(\mathcal{D}\), where:
\begin{itemize}[noitemsep,left=0pt]
        \item \(x\): The input prompt or context.
        \item \(y_w\): better personas, corresponding to more optimal update directions.
        \item \(y_l\): poor personas, corresponding to less effective update directions.
\end{itemize}

    A scaling factor that controls the sensitivity of the loss function to the difference between the preferred and less preferred outputs. Larger values emphasize the contrast between the two.
    The conditional probabilities of the preferred (\(y_w\)) and less preferred (\(y_l\)) outcomes under the current model.
    The conditional probabilities of the preferred and less preferred outcomes under the reference model.
    The loss is computed as an average over the entire dataset \(\mathcal{D}\), which contains human-annotated preference pairs \((x, y_w, y_l)\).

\paragraph{SFT Loss}
The SFT loss is used to aline the model output with high-quality refined-persona candidates. The loss is computed as the negative log-likelihood of the reference outputs:
\[
\mathcal{L}_{\text{SFT}}(\pi_\theta) = - \mathbb{E}_{(x, y_w) \sim \mathcal{D}} \big[ \log \pi_\theta(y_w|x) \big]. \tag{12}
\]
where \(\mathcal{D}\) is the dataset of supervised examples, \(x\) represents the input context, and \(y\) is the corresponding ground truth persona refinement.

\paragraph{Combined Loss for Iterative Training}
To achieve robust refinement across iterations, we combine the SFT loss and DPO loss :
\[
\mathcal{L}(\pi_\theta; \pi_{\text{ref}}) = \mathcal{L}_{\text{DPO}}(\pi_\theta; \pi_{\text{ref}}) +  \mathcal{L}_{\text{SFT}}(\pi_\theta). \tag{13}
\]

\clearpage
\section{Dynamic Persona Modeling Task}

In the main experiment, we focused on the dynamic persona modeling task, where different methods are employed to perform four rounds of updates on the test set. These iterative updates provided specific values for predicting future user behavior, enabling us to assess the accuracy and effectiveness of each method in forecasting user actions. By evaluating the Mean Absolute Error (MAE) across various domains before and after refinement, we are able to determine the improvements achieved through each method. The results, detailed in Table \ref{tab:Deeper_results},Table \ref{tab:fullregen_results},Table \ref{tab:slideregen_results},Table \ref{tab:incupdate_results},Table \ref{tab:hiermerge_results}, highlight the performance gains and validate the superiority of the proposed approaches in enhancing prediction accuracy.

\begin{table}[h]
    \centering
    \scriptsize
    \def\arraystretch{.99}
    \setlength{\tabcolsep}{0.42em}
    \begin{tabular}{lccccc}
      \toprule
      \textbf{Domain} & $\varepsilon_{1|\mathcal{S}_0}$ & $\varepsilon_{2|\mathcal{S}_1}$ & $\varepsilon_{3|\mathcal{S}_2}$ & $\varepsilon_{4|\mathcal{S}_3}$ & $\varepsilon_{5|\mathcal{S}_4}$ \\
      \midrule
      Art Crafts and Sewing     & 0.76 & 0.43 & 0.44 & 0.41 & 0.40 \\
      Automative                & 0.84 & 0.61 & 0.60 & 0.59 & 0.57 \\
      Book                      & 1.00 & 0.79 & 0.70 & 0.68 & 0.67 \\
      Clothing Shoes and Jewelry& 1.00 & 0.88 & 0.84 & 0.75 & 0.74 \\
      Grocery and Gourmet Food  & 1.19 & 0.89 & 0.79 & 0.77 & 0.73 \\
      Local Business            & 1.04 & 0.80 & 0.70 & 0.70 & 0.69 \\
      Movie                     & 0.87 & 0.73 & 0.73 & 0.72 & 0.72 \\
      Movies and TV             & 1.12 & 0.98 & 0.83 & 0.79 & 0.77 \\
      Recipe                    & 0.91 & 0.72 & 0.65 & 0.59 & 0.58 \\
      Sports and Outdoors       & 0.91 & 0.80 & 0.68 & 0.66 & 0.66 \\
      \bottomrule
    \end{tabular}
    \caption{\method Results}
    \label{tab:Deeper_results}
    \vspace{-0.1cm}
  \end{table}

\begin{table}[h]
    \centering
    \scriptsize
    \def\arraystretch{.99}
    \setlength{\tabcolsep}{0.42em}
    \begin{tabular}{lccccc}
      \toprule
      \textbf{Domain} & $\varepsilon_{1|\mathcal{S}_0}$ & $\varepsilon_{2|\mathcal{S}_1}$ & $\varepsilon_{3|\mathcal{S}_2}$ & $\varepsilon_{4|\mathcal{S}_3}$ & $\varepsilon_{5|\mathcal{S}_4}$ \\
      \midrule
      Art Crafts and Sewing     & 0.76 & 0.77 & 0.74 & 0.69 & 0.75 \\
      Automative                & 0.84 & 0.92 & 0.92 & 0.92 & 0.93 \\
      Book                      & 1.00 & 1.03 & 1.00 & 1.00 & 1.00 \\
      Clothing Shoes and Jewelry& 1.00 & 1.03 & 1.03 & 1.03 & 1.03 \\
      Grocery and Gourmet Food  & 1.19 & 1.20 & 1.16 & 1.14 & 1.13 \\
      Local Business            & 1.04 & 1.04 & 1.02 & 1.03 & 1.02 \\
      Movie                     & 0.87 & 0.83 & 0.84 & 0.84 & 0.85 \\
      Movies and TV             & 1.12 & 1.12 & 1.13 & 1.13 & 1.11 \\
      Recipe                    & 0.91 & 0.92 & 0.83 & 0.84 & 0.85 \\
      Sports and Outdoors       & 0.91 & 0.93 & 0.85 & 0.88 & 0.88 \\
      \bottomrule
    \end{tabular}
    \caption{FullRegen (GPT-4o-mini) Results}
    \label{tab:fullregen_results}
    \vspace{-0.1cm}
  \end{table}

\begin{table}[h]
    \centering
    \scriptsize
    \def\arraystretch{.99}
    \setlength{\tabcolsep}{0.42em}
    \begin{tabular}{lccccc}
      \toprule
      \textbf{Domain} & $\varepsilon_{1|\mathcal{S}_0}$ & $\varepsilon_{2|\mathcal{S}_1}$ & $\varepsilon_{3|\mathcal{S}_2}$ & $\varepsilon_{4|\mathcal{S}_3}$ & $\varepsilon_{5|\mathcal{S}_4}$ \\
      \midrule
      Art Crafts and Sewing     & 0.76 & 0.81 & 0.72 & 0.66 & 0.74 \\
      Automative                & 0.84 & 0.96 & 0.91 & 0.91 & 0.90 \\
      Book                      & 1.00 & 1.06 & 1.02 & 1.03 & 1.02 \\
      Clothing Shoes and Jewelry& 1.00 & 1.09 & 1.08 & 1.02 & 1.07 \\
      Grocery and Gourmet Food  & 1.19 & 1.14 & 1.13 & 1.13 & 1.09 \\
      Local Business            & 1.04 & 1.06 & 1.05 & 1.03 & 1.03 \\
      Movie                     & 0.87 & 0.84 & 0.85 & 0.87 & 0.86 \\
      Movies and TV             & 1.12 & 1.14 & 1.16 & 1.17 & 1.14 \\
      Recipe                    & 0.91 & 0.92 & 0.89 & 0.87 & 0.87 \\
      Sports and Outdoors       & 0.91 & 0.94 & 0.87 & 0.92 & 0.93 \\
      \bottomrule
    \end{tabular}
    \caption{SlideRegen (GPT-4o-mini) Results}
    \label{tab:slideregen_results}
    \vspace{-0.1cm}
  \end{table}

\begin{table}[h]
    \centering
    \scriptsize
    \def\arraystretch{.99}
    \setlength{\tabcolsep}{0.42em}
    \begin{tabular}{lccccc}
      \toprule
      \textbf{Domain} & $\varepsilon_{1|\mathcal{S}_0}$ & $\varepsilon_{2|\mathcal{S}_1}$ & $\varepsilon_{3|\mathcal{S}_2}$ & $\varepsilon_{4|\mathcal{S}_3}$ & $\varepsilon_{5|\mathcal{S}_4}$ \\
      \midrule
      Art Crafts and Sewing     & 0.76 & 0.71 & 0.70 & 0.59 & 0.66 \\
      Automative                & 0.84 & 0.88 & 0.81 & 0.87 & 0.82 \\
      Book                      & 1.00 & 0.96 & 0.94 & 0.92 & 0.93 \\
      Clothing Shoes and Jewelry & 1.00 & 1.00 & 0.97 & 0.93 & 0.92 \\
      Grocery and Gourmet Food  & 1.19 & 1.10 & 1.05 & 1.05 & 1.04 \\
      Local Business            & 1.04 & 0.97 & 0.94 & 0.93 & 0.91 \\
      Movie                     & 0.87 & 0.76 & 0.76 & 0.76 & 0.75 \\
      Movies and TV             & 1.12 & 1.06 & 1.06 & 1.04 & 1.05 \\
      Recipe                    & 0.91 & 0.91 & 0.84 & 0.84 & 0.81 \\
      Sports and Outdoors       & 0.91 & 0.89 & 0.85 & 0.85 & 0.83 \\
      \bottomrule
    \end{tabular}
    \caption{IncUpdate (GPT-4o-mini) Results}
    \label{tab:incupdate_results}
   \end{table}
  
\newpage

\begin{table}[t]
    \centering
    \scriptsize
    \def\arraystretch{.99}
    \setlength{\tabcolsep}{0.42em}
    \begin{tabular}{lccccc}
      \toprule
      \textbf{Domain} & $\varepsilon_{1|\mathcal{S}_0}$ & $\varepsilon_{2|\mathcal{S}_1}$ & $\varepsilon_{3|\mathcal{S}_2}$ & $\varepsilon_{4|\mathcal{S}_3}$ & $\varepsilon_{5|\mathcal{S}_4}$ \\
      \midrule
      Art Crafts and Sewing     & 0.76 & 0.75 & 0.73 & 0.67 & 0.72 \\
      Automative                & 0.84 & 0.88 & 0.86 & 0.94 & 0.85 \\
      Book                      & 1.00 & 1.03 & 0.98 & 0.97 & 0.93 \\
      Clothing Shoes and Jewelry& 1.00 & 1.04 & 1.01 & 1.04 & 1.04 \\
      Grocery and Gourmet Food  & 1.19 & 1.19 & 1.16 & 1.18 & 1.14 \\
      Local Business            & 1.04 & 1.04 & 1.00 & 1.01 & 1.00 \\
      Movie                     & 0.87 & 0.82 & 0.83 & 0.83 & 0.82 \\
      Movies and TV             & 1.12 & 1.12 & 1.11 & 1.10 & 1.08 \\
      Recipe                    & 0.91 & 0.94 & 0.87 & 0.81 & 0.81 \\
      Sports and Outdoors       & 0.91 & 0.90 & 0.89 & 0.91 & 0.87 \\
      \bottomrule
    \end{tabular}
    \caption{HierMerge (GPT-4o-mini) Results}
    \label{tab:hiermerge_results}

  \end{table}

\clearpage
\section{What Enables \method's Effectiveness}
Below, we present an in-depth analysis of the mechanisms underlying \method’s effectiveness.

\subsection{Proving Effectiveness of Direction search}
Firstly, we prove the effectiveness of the direction search method by comparing its performance with a direct refinement using the frozen model(GPT-4o-mini and the base model, Llama3.1-8b-Instruct), through a single round of refinement. The details of the experimental results are as follows Label \ref{tab:direction_search_use}.

\begin{itemize}[noitemsep,left=0pt]
    \item \textbf{Baseline 1} Directly Refine personas with the base model (Llama3.1-8b-Instruct) 
    \item \textbf{Baseline 2} Directly Refine personas with the more powerful model (GPT-4o-mini) 
    \item \textbf{\method} Refine personas with auto-direction search mechanism
\end{itemize}

\begin{table}[h]
\centering
\scriptsize
\def\arraystretch{.99}
\setlength{\tabcolsep}{0.42em}
  \begin{tabular}{lcccc}
    \toprule
    & \multicolumn{1}{c}{\textbf{\begin{tabular}[c]{@{}c@{}}Before \\ Update\end{tabular}}} & \multicolumn{3}{c}{\textbf{\begin{tabular}[c]{@{}c@{}}After \\ Update\end{tabular}}} \\
    \midrule
    \textbf{Domain} & $\varepsilon_{1|\mathcal{S}_0}$ & \begin{tabular}[c]{@{}c@{}}$\varepsilon_{2|\mathcal{S}_1}$ \\ (\method)\end{tabular}& \begin{tabular}[c]{@{}c@{}}$\varepsilon_{2|\mathcal{S}_1}$ \\(GPT-4o-mini)\end{tabular} & \begin{tabular}[c]{@{}c@{}}$\varepsilon_{2|\mathcal{S}_1}$ \\ (Llama3.1-8b-Instruct)\end{tabular}\\
    \midrule
    A                    & 0.91 & 0.72 & 0.99 & 1.07 \\
    B                      & 1.01 & 0.79 & 1.20 & 1.21 \\
    C& 1.03 & 0.88 & 1.09 & 1.14 \\
    D            & 1.04 & 0.80 & 1.20 & 1.19 \\
    E             & 1.18 & 0.98 & 1.24 & 1.19 \\
    F                     & 0.85 & 0.73 & 0.84 & 0.87 \\
    G     & 0.74 & 0.43 & 0.95 & 0.89 \\
    H                & 0.85 & 0.61 & 1.02 & 1.08 \\
    I  & 1.26 & 0.89 & 1.26 & 1.19 \\
    J       & 0.96 & 0.80 & 1.04 & 1.02 \\
    \bottomrule
  \end{tabular}
  \caption{Future behaviour prediction errors before and after one-step refinement with \method and the frozen models.}
  \label{tab:direction_search_use}
  \vspace{-0.1cm}
\end{table}

\subsection{Proving the Effectiveness of Balanced Reward}

In this analysis, we aim to demonstrate the effectiveness of the balanced reward strategy by comparing it against two baseline reward settings. Specifically, we evaluate how different reward configurations influence the performance of the model during the refinement process.

\paragraph{Baseline Reward Settings}
We establish two baseline configurations to assess the impact of reward settings:

\begin{itemize}[noitemsep,left=0pt]
    \item \textbf{Baseline 1: Future Advancement Only} \\
    In this setting, the reward at each timestep $t$ is solely based on future advancement. Mathematically, this is defined as:
    \[
    r_t = r_{\text{fut}} = r_t^{\text{fut}}. \tag{14}
    \]
    
    \item \textbf{Baseline 2: Decayed Rewards} \\
    Here, we incorporate past, current, and future rewards with decay factors applied to past and current rewards. The reward at timestep $t$ is calculated as:
    \[
    r_t = r_{\text{decay}} = 0.25 \cdot r_t^{\text{prev}} + 0.5 \cdot r_t^{\text{curr}} + r_t^{\text{fut}}. \tag{15}
    \]
    where the decay factor $y = 0.5$ is applied to both past and current rewards.
\end{itemize}

\paragraph{Our Reward Setting: Balanced Rewards}
Our proposed reward setting balances the three components—past, current, and future—without applying decay factors. The reward at timestep $t$ is defined as:
\[
r_t = r_t^{\text{prev}} + r_t^{\text{curr}} + r_t^{\text{fut}}. \tag{16}
\]
This approach ensures that all three goals are equally considered during the refinement process.

\textbf{Experimental Results}
The experimental results comparing the baseline reward settings with our balanced reward strategy are presented in Table \ref{tab:balanced_reward_results_refined}, which showcase future prediction errors across various domains before and after one-step refinement under different reward configurations.

\begin{table}[h]
\centering
\scriptsize
\def\arraystretch{.99}
\setlength{\tabcolsep}{0.42em}
  \begin{tabular}{lcccc}
    \toprule
    & \multicolumn{1}{c}{\textbf{\begin{tabular}[c]{@{}c@{}}Before \\ Update\end{tabular}}} & \multicolumn{3}{c}{\textbf{\begin{tabular}[c]{@{}c@{}}After \\ Update\end{tabular}}} \\
    \midrule
    \textbf{Domain} & $\varepsilon_{1|\mathcal{S}_0}$ & \begin{tabular}[c]{@{}c@{}}$\varepsilon_{2|\mathcal{S}_1}$ \\ (\method)\end{tabular}& \begin{tabular}[c]{@{}c@{}}$\varepsilon_{2|\mathcal{S}_1}$ \\($r_{\text{fut}}$)\end{tabular} & \begin{tabular}[c]{@{}c@{}}$\varepsilon_{2|\mathcal{S}_1}$ \\ ($r_{\text{decay}}$)\end{tabular}\\
    \midrule
    A                    & 0.91 & 0.72 & 0.81 & 0.74 \\
    B                      & 1.01 & 0.79 & 0.86 & 0.84 \\
    C & 1.03 & 0.88 & 0.95 & 0.92 \\
    D            & 1.04 & 0.80 & 0.88 & 0.84 \\
    E             & 1.18 & 0.98 & 1.03 & 1.03 \\
    F                     & 0.85 & 0.73 & 0.81 & 0.77 \\
    G     & 0.74 & 0.43 & 0.59 & 0.51 \\
    H                & 0.85 & 0.61 & 0.83 & 0.68 \\
    I  & 1.26 & 0.89 & 0.94 & 0.92 \\
    J       & 0.96 & 0.80 & 0.85 & 0.83 \\
    \bottomrule
  \end{tabular}
  \caption{Balanced reward (\method) vs. Baseline reward settings results}
  \label{tab:balanced_reward_results_refined}
  \vspace{-0.1cm}
\end{table}

\subsection{Proving the Effectiveness of Iterative RL Training}

In this analysis, we aim to demonstrate the effectiveness of iterative offline RL training by comparing the performance of the model after one iteration of refinement versus two iterations. This comparison helps to understand whether additional refinement iterations contribute to improved model performance.

\textbf{Baseline} To evaluate the impact of iterative RL training, we establish two baseline configurations:
\begin{itemize}[noitemsep,left=0pt]
    \item \textbf{Baseline 1: Single Iteration} \\
    The model undergoes one iteration of training.
    
    \item \textbf{Baseline 2: Two Iterations} \\
   The model undergoes two consecutive iterations of training to assess whether additional refinement leads to further performance gains.
\end{itemize}

\textbf{Experimental Results} The experimental results comparing single and double iterations of refinement are presented in Table \ref{tab:iterative_rl_training_results_refined}.

\begin{table}[h]
\centering
\scriptsize
\def\arraystretch{.99}
\setlength{\tabcolsep}{0.42em}
  \begin{tabular}{lccc}
    \toprule
    & \multicolumn{1}{c}{\textbf{\begin{tabular}[c]{@{}c@{}}Before \\ Update\end{tabular}}} & \multicolumn{2}{c}{\textbf{\begin{tabular}[c]{@{}c@{}}After \\ Update\end{tabular}}} \\
    \midrule
    \textbf{Domain} & $\varepsilon_{1|\mathcal{S}_0}$ & \begin{tabular}[c]{@{}c@{}}$\varepsilon_{2|\mathcal{S}_1}$ \\ (\method(Iter1))\end{tabular}& \begin{tabular}[c]{@{}c@{}}$\varepsilon_{2|\mathcal{S}_1}$ \\(\method(Iter2))\end{tabular} \\
    \midrule
    A                    & 0.91 & 0.84 & 0.72 \\
    B                      & 1.01 & 0.93 & 0.79 \\
    C & 1.03 & 0.99 & 0.88 \\
    D            & 1.04 & 0.98 & 0.80 \\
    E             & 1.18 & 1.12 & 0.98 \\
    F                     & 0.85 & 0.78 & 0.73 \\
    G     & 0.74 & 0.65 & 0.43 \\
    H                & 0.85 & 0.87 & 0.61 \\
    I  & 1.26 & 1.03 & 0.89 \\
    J       & 0.96 & 0.91 & 0.80 \\
    \bottomrule
  \end{tabular}
  \caption{Iterative RL Training (\method) Results comparison: Single vs. Double Training Iterations}
  \label{tab:iterative_rl_training_results_refined}
  \vspace{-0.1cm}
\end{table}

\newpage
\subsection{Proving the Effectiveness of Introducing Prediction Results in Refinement Paradigm}

In this analysis, we aim to demonstrate the effectiveness of incorporating prediction results into the paradigm. Specifically, we leverage the iterative RL training framework of \method to enhance IncUpdate(Inc-FT), which is the best performing baseline and based on paradigm of Persona Extension. This enable auto-direction search in traditional dynamic persona paradigm which does not involve prediction results into observations. The comparison between \method and Inc-FT helps to understand whether integrating prediction results helps direction search.

\textbf{Experimental Results} The experimental results are presented in Table \ref{tab:prediction_result_effectiveness_results_refined}. 

\begin{table}[h]
\centering
\scriptsize
\def\arraystretch{.99}
\setlength{\tabcolsep}{0.42em}
  \begin{tabular}{lccc}
    \toprule
    & \multicolumn{1}{c}{\textbf{\begin{tabular}[c]{@{}c@{}}Before \\ Update\end{tabular}}} & \multicolumn{2}{c}{\textbf{\begin{tabular}[c]{@{}c@{}}After \\ Update\end{tabular}}} \\
    \midrule
    \textbf{Domain} & $\varepsilon_{1|\mathcal{S}_0}$ & \begin{tabular}[c]{@{}c@{}}$\varepsilon_{2|\mathcal{S}_1}$ \\ (\method)\end{tabular}& \begin{tabular}[c]{@{}c@{}}$\varepsilon_{2|\mathcal{S}_1}$ \\(Inc-FT))\end{tabular} \\
    \midrule
    Recipe                    & 0.91 & 0.72 & 0.77 \\
    Book                      & 1.01 & 0.79 & 0.85 \\
    Clothing Shoes and Jewelry& 1.03 & 0.88 & 0.95 \\
    Local Business            & 1.04 & 0.80 & 0.81 \\
    Movies and TV             & 1.18 & 0.98 & 1.02 \\
    Movie                     & 0.85 & 0.73 & 0.79 \\
    Art Crafts and Sewing     & 0.74 & 0.43 & 0.62 \\
    Automative                & 0.85 & 0.61 & 0.78 \\
    Grocery and Gourmet Food  & 1.26 & 0.89 & 0.97 \\
    Sports and Outdoors       & 0.96 & 0.80 & 0.81 \\
    \bottomrule
  \end{tabular}
  \caption{Effectiveness of Introducing Prediction Results: \method vs. IncUpdate (Inc-FT)}
  \label{tab:prediction_result_effectiveness_results_refined}
  \vspace{-0.1cm}
\end{table}

\begin{table*}[t]
      \small
      \centering
      \caption{Important Profiling Dimensions in the Book Domain}
      \begin{tabular}{lp{7.5cm}}  
      \toprule
      \textbf{Dimensions}            & \textbf{High-Frequency Terms and Frequency}                                                                                                      \\ \midrule
      \textbf{Story \& Plot}       & story (759), experience (596), reader (579), narrative (566), storytelling (445), development (440), plot (286), adventure (196), fantasy (187), suspense (156), mystery (155), action (150), passion (149), thriller (109), journey (92), arc (75), drama (66), protagonist (58), redemption (54)           \\ \midrule
      \textbf{Genre \& Theme}      & theme (655), genre (647), romance (388), aspect (406), level (379), content (329), complexity (325), depth (325), world (277), novel (212), topic (189), element (231), idea (189), nuance (148), literature (137), nature (118), issue (122), setting (88), balance (97), thought (96)      \\ \midrule
      \textbf{Author \& Character} & author (143), quality (155), character (600), characteristic (93), identity (70), protagonist (58) \\ \midrule
      \textbf{Emotion \& Experience}  & affinity (217), appreciation (700), experience (596), willingness (685), desire (563), love (410), enthusiasm (377), resonance (352), emotion (185), escapism (203), curiosity (182), expectation (167), favor (153), enjoyment (136), excitement (121), comfort (117) \\ \midrule
      \textbf{User Behavior Traits}  & range (568), star (518), growth (318), perspective (279), engagement (250), title (247), tendency (185), habit (155), exploration (140), investment (86), variety (67) \\ \midrule
      \textbf{User Personality \& Values} & willingness (685), preference (581), individual (388), value (221), self (183), discerning (183), personality (132), creativity (94), empathy (85), adaptability (90) \\ \midrule
      \textbf{Social \& Cultural Context} & life (198), community (164), time (181), boundary (99), need (112), culture (77), knowledge (88), learning (95), justice (61) \\ \midrule
      \textbf{Relationship \& Connection} & connection (474), relationship (411), choice (98), family (67), interaction (57) \\ \bottomrule
      \end{tabular}
      \label{tab:profiling_dimensions}
    \end{table*}
    
\newpage
\section{Persona Probing}

\subsection{Important Profiling Dimensions in Book Domain}
    Table ~\ref{tab:profiling_dimensions} summarizes the key profiling dimensions for users in the book domain, along with the high-frequency terms and their frequencies within each dimension. These dimensions include “Story \& Plot,” “Genre \& Theme,” “Author \& Character,” among others, which encapsulate critical aspects of user preferences and behaviors. The table highlights the most commonly used terms, such as “story,” “experience,” and “reader” under the “Story \& Plot” dimension, providing insights into what users value when engaging with book-related content.

\subsection{Insights into User Group Characteristics}
Table ~\ref{tab:group_user} illustrates the unique adjectives frequently associated with specific user groups, providing a detailed view of the preferences that distinguish these groups. For instance, Group 1 exhibits traits such as “romantic” and “dedicated,” while Group 4 emphasizes “practical” and “cultural” preferences. These findings underscore the variation in user characteristics, enabling targeted persona optimization based on group-specific attributes.

\begin{table}[h]
  \small
  \centering
  \caption{Group-level preference for users}
  \begin{tabular}{lp{5cm}} 
  \toprule
  \textbf{User Groups}    & \textbf{Unique High-Frequency Adjectives}  \\
  \midrule
  Group 1    & romantic, paranormal, voracious, \\  
             & dedicated, afraid                                    \\ \midrule
  Group 2    & notable, humorous, unconventional, \\  
             & close, entertaining                                \\ \midrule
  Group 3    & dramatic, dedicated, \\   
             & resonant                                                         \\ \midrule
  Group 4    & practical, spiritual, historical, cultural, \\ 
             & likely, playful, dynamic, inspirational, close \\ \midrule
  Group 5    & thoughtful, non, suspenseful, \\   
             & fiction, dynamic, engaging, immersive                   \\ 
  \bottomrule
  \end{tabular}
  \label{tab:group_user}
\end{table}
      
\clearpage
\section{Additional Analysis}
\subsection{Baselines Enhanced with Stronger Backbones and Task-Specific SFT}
To further validate the limitations of existing extension- and regeneration-based methods in dynamic persona modeling, we conduct additional experiments under two enhancement conditions:

\begin{itemize}
  \item \textbf{(1) Stronger Backbone:} All baselines are implemented with GPT-4o-mini as the underlying predictor and persona generator.
  \item \textbf{(2) Task-Specific SFT:} We fine-tune baselines using task-specific supervision constructed via self-sampling and reject sampling, based on future prediction error comparisons.
\end{itemize}

In both settings, we follow the same multi-round dynamic persona modeling task described in the main paper. Table~\ref{tab:enhanced_baseline} summarizes the results after the first persona update.

\begin{table}[h]
\scriptsize
\centering
\caption{Comparison of MAE before and after the first update across methods with strong backbones or task-specific SFT.}
\label{tab:enhanced_baseline}
\begin{tabular}{lccc}
\toprule
\textbf{Method} & \textbf{Before Update} & \textbf{After Update} & \textbf{MAE Reduction} \\
\midrule
\textbf{\method} & 0.964 & \textbf{0.763} & \textbf{+20.9\%} \\
\cdashlinelr{1-4}
\multicolumn{4}{>{\columncolor{gray!4}}c}
{\textit{Stronger Backbone: GPT-4o-mini}} \\
HierMerge & 0.964 & 0.974 & -1.0\% \\
IncUpdate  & 0.964 & 0.924 & +4.1\% \\
SlideRegen & 0.964 & 0.996 & -3.3\% \\
FullRegen & 0.964 & 0.979 & -1.6\% \\
\cdashlinelr{1-4}
\multicolumn{4}{>{\columncolor{gray!4}}c}
{\textit{Task-Specific SFT: Llama-3.1-8B-Instruct}} \\
HierMerge & 0.964 & 0.985 & -2.2\% \\
IncUpdate  & 0.964 & 0.963 & +0.1\% \\
SlideRegen  & 0.964 & 1.025 & -6.3\% \\
FullRegen  & 0.964 & 0.992 & -2.9\% \\
\bottomrule
\end{tabular}
\end{table}

As shown above, even when enhanced with stronger backbones or supervised fine-tuning, baseline methods fail to achieve meaningful optimization. In contrast, \method achieves a substantial MAE reduction. This provides further evidence that existing approaches suffer from a fundamental gap between persona updating and optimization, which \method effectively bridges through reward-guided direction refinement.

\subsection{Reusability and Scalability of \method Framework}
Although \method adopts a reinforcement learning-based training pipeline, which involves moderate computational overhead, we emphasize that its resulting dataset is reusable, model-agnostic, and adaptable to various downstream tasks. This significantly reduces the long-term costs of applying dynamic persona modeling in practice.

\paragraph{1. Domain Generalization.}  
The \method dataset is constructed using six training domains and generalizes effectively to unseen settings. As shown in the main results, \method achieves an average 36.4\% MAE reduction across four unseen domains after four update rounds.

\paragraph{2. Task Generalization.}  
To evaluate task transferability, we construct a pair-wise prediction test on 600 users. After one refinement step, \method improves choice accuracy from 61\% to 68\%, reflecting an 11.4\% increase in task-specific accuracy.

\begin{table}[h]
\scriptsize
\centering
\caption{Pair-wise choice accuracy before and after refinement.}
\label{tab:pairwise}
\begin{tabular}{lccc}
\toprule
\textbf{Test Set} & \textbf{Before} & \textbf{After} & \textbf{Accuracy Increase} \\
\midrule
600 Users & 0.61 & 0.68 & +11.4\% \\
\bottomrule
\end{tabular}
\end{table}

\paragraph{3. Model Generalization.}  
We further apply the \method dataset to fine-tune a different backbone, Qwen2-7B-Instruct~\cite{qwen2}. The results demonstrate consistent gains: 13.1\% MAE reduction in seen domains and 10.8\% in unseen domains.

\begin{table}[h]
\scriptsize
\centering
\caption{Performance of Qwen2-7B-Instruct fine-tuned on \method dataset.}
\label{tab:qwen2}
\begin{tabular}{lccc}
\toprule
\textbf{Domain Type} & \textbf{Before MAE} & \textbf{After MAE} & \textbf{MAE Reduction} \\
\midrule
Seen Domain & 0.99 & 0.86 & 13.1\% \\
Unseen Domain & 0.925 & 0.825 & 10.8\% \\
\bottomrule
\end{tabular}
\end{table}

\paragraph{Conclusion.}  
While \method's offline RL training incurs moderate one-time cost, its dataset offers strong generalization across domains, tasks, and models—eliminating the need for repeated reward computation or full-model fine-tuning. This makes \method a practical and scalable solution for real-world dynamic persona modeling.

\clearpage
\section{Case Study: A User in Book Domain}
To deeply evaluate the performance of different persona updates methods for dynamic persona modeling, we selected a single user from the book domain. This domain provides a complex and rich context, as users often demonstrate evolving preferences, diverse genre interests, and emotional connections with books over time. In case study, we focus on the improvements of future prediction task over four update rounds and the evolution of user's personas with \method as the update method.
\subsection{Dynamic Persona Modeling}
We first compares five persona update methods: \method, FullRegen, SlideRegen, IncUpdate, and HierMerge, focusing on the evolution of the user’s persona across 4 update steps and evaluate their effectiveness based on the future prediction error (MAE) at each step, as shown in Table \ref{tab:future_prediction_error}. 
 The results of this case demonstrates that \method consistently reduces prediction error across refinement steps, achieving continual persona optimization, while all baseline methods not only fail to improve but also degrade persona quality over time.

    \begin{table}[!h]
      \small
      \centering
      \setlength{\tabcolsep}{0.8em}
      \caption{Future prediction error across 4 persona updates.}
      \begin{tabular}{lccccc}
        \toprule
        \textbf{Domain} & $\varepsilon_{1|\mathcal{S}_0}$ & $\varepsilon_{2|\mathcal{S}_1}$ & $\varepsilon_{3|\mathcal{S}_2}$& $\varepsilon_{4|\mathcal{S}_3}$& $\varepsilon_{5|\mathcal{S}_4}$ \\
        \midrule
        \method    & 1.1 & 1.0 & 0.5 & 0.3 & 0.2 \\
        FullRegen & 1.1 & 1.6 & 1.5 & 1.5 & 1.7 \\
        SlideRegen & 1.1 & 1.4 & 1.8 & 1.9 & 1.5 \\
        IncUpdate & 1.1 & 1.3 & 1.6 & 1.7 & 1.2 \\
        HierMerge & 1.1 & 1.7 & 1.8 & 1.9 & 1.3 \\ 
        \bottomrule
      \end{tabular}
      \label{tab:future_prediction_error}
    \end{table}

\newpage
\subsection{Persona Evolution}
In this section, we display the evolution of the user’s persona and corresponding observations in each window over four persona update rounds with \method (Table ~\ref{tab:persona_evolve_1}, Table ~\ref{tab:persona_evolve_2}, Table ~\ref{tab:persona_evolve_3}, Table ~\ref{tab:persona_evolve_4}, Table ~\ref{tab:persona_evolve_5}).
\begin{table*}[h!]
  \centering
  \small
  \resizebox{\linewidth}{!}{
  \begin{tabular}{ll}
  \toprule
  \multicolumn{2}{c}{\textbf{Examples of Persona Evolution of \method(\textit{A User from Book Domain})}} \\ \midrule
  \multicolumn{1}{l}{$\mathcal{S}_0$ and observartions in  $\mathcal{W}_1$} &
  \parbox{0.75\textwidth}{"The user appears to be an individual with a strong appreciation for both creative expression and practical knowledge. Their high ratings for works that blend storytelling with deeper themes suggest a reflective and imaginative personality. They likely enjoy narratives that not only entertain but also provoke thought and inspire personal growth. The consistent high ratings for books related to finance and funding indicate a keen interest in practical skills and knowledge, suggesting that they may be entrepreneurial or involved in creative industries where funding and financial acumen are crucial. This duality in their reading preferences points to a well-rounded individual who values both the arts and the practicalities of life. The user’s enjoyment of stories that involve mystery and justice indicates a fascination with problem-solving and perhaps a desire for moral clarity in complex situations. Their varied interests suggest they are open-minded and curious, likely seeking to expand their horizons through diverse genres. Overall, this user embodies a blend of creativity and pragmatism, with a penchant for narratives that inspire, educate, and entertain, reflecting a holistic approach to personal and professional development."} \\ \cmidrule{2-2} 
  \multicolumn{1}{l}{} &
  \parbox{0.75\textwidth}{item\_list = ["Insights of an Ordinary Man","Aura Advantage: How the Colors in Your Aura Can Help You Attain What You Desire and Attract Success","Georgia Bound", "Fifth Life of the Cat Woman", "Lady Justice and the Assassin", "Lord James Harrington and the Winter Mystery (Volume 1)", "Elemental Magic", "Enchantments of the Heart","The Grimm Diaries Prequels 1-6", "The Quest for the Crystal Skulls"]\\\\
  actual\_ratings =   [5, 5, 5, 5, 5, 5, 4, 3, 5, 5]\\
  predict\_ratings = [4, 3, 3, 4, 5, 4, 3, 3, 3, 4]} 
  \\\bottomrule
    \end{tabular}
    }
    \caption{Examples of $\mathcal{S}_0$ and observartions in $\mathcal{W}_1$}
    \label{tab:persona_evolve_1}
    \end{table*}

\begin{table*}[h!]
  \centering
  \small
  \resizebox{\linewidth}{!}{
  \begin{tabular}{ll}
  \toprule
  \multicolumn{2}{c}{\textbf{Examples of Persona Evolution of \method(\textit{A User from Book Domain})}} \\ \midrule
  \multicolumn{1}{l}{$\mathcal{S}_1$ and observartions in  $\mathcal{W}_2$} &
  \parbox{0.75\textwidth}{"The user appears to be a creative and open-minded individual with a strong appreciation for diverse forms of expression. Their high ratings for a wide range of books, including those that blend fiction with spiritual or self-help themes, suggest a deep interest in personal growth and self-discovery. This inclination towards inspirational and motivational content may indicate that the user is seeking guidance or validation in their own life journey. Their consistent 5-star ratings across various genres, including mystery and fantasy, demonstrate a willingness to explore new ideas and perspectives. The user's appreciation for practical knowledge and skills, as evident in their interest in finance and funding, highlights their pragmatic side and potential entrepreneurial or creative pursuits. Their enjoyment of stories that involve justice and problem-solving suggests a strong sense of morality and a desire for fairness. Overall, this user embodies a unique blend of creativity, curiosity, and practicality, with a penchant for narratives that inspire, educate, and entertain. They seem to value diversity and are not afraid to explore unconventional ideas, making them a highly open-minded and adaptable individual."} \\ \cmidrule{2-2} 
  \multicolumn{1}{l}{} &
  \parbox{0.75\textwidth}{item\_list= [ "A Hidden Fire: Elemental Mysteries Book 1", "It's Just a Little Crush: A Lizzie Hart Mystery (The Lizzie Hart Mysteries) (Volume 1)","The Devil Drinks Coffee", "The Depot: When Life and Death Cross Tracks","Lady Justice and the Lottery (Volume 14)", "Wuthering Heights (Penguin Classics)","The Ambitious Card (An Eli Marks Mystery) (Volume 1)","Miracles \& Mischief: Meringue for your Brain", "The Color of Greed: Raja Williams Series (Raja Williams Mystery Thriller Series)", "Thursday Morning Breakfast (and Murder) Club"]\\\\
actual\_ratings= [5, 5, 5, 5, 5, 5, 5, 5, 5, 5]\\
predict\_ratings = [4, 4, 3, 4, 5, 5, 4, 3, 4, 4]} \\\bottomrule
  \end{tabular}
  }
  \caption{Examples of $\mathcal{S}_1$ and observartions in $\mathcal{W}_2$.}
  \label{tab:persona_evolve_2}
  \end{table*}

  \newpage

  \begin{table*}[h!]
    \centering
    \small
    \resizebox{\linewidth}{!}{
    \begin{tabular}{ll}
    \toprule
    \multicolumn{2}{c}{\textbf{Examples of Persona Evolution of \method(\textit{A User from Book Domain})}} \\ \midrule
   
    \multicolumn{1}{l}{$\mathcal{S}_2$ and observartions in  $\mathcal{W}_3$} &
    \parbox{0.75\textwidth}{ "The user appears to be an extremely enthusiastic and optimistic individual who thoroughly enjoys a wide range of books across various genres. Their high ratings for a vast array of titles, including those that blend fiction with spiritual or self-help themes, suggest a deep interest in personal growth and self-discovery. This inclination towards inspirational and motivational content may indicate that the user is seeking guidance or validation in their own life journey. Their consistent 5-star ratings across various genres, including mystery and fantasy, demonstrate a willingness to explore new ideas and perspectives. The user's appreciation for practical knowledge and skills, as evident in their interest in finance and funding, highlights their pragmatic side and potential entrepreneurial or creative pursuits. Their enjoyment of stories that involve justice and problem-solving suggests a strong sense of morality and a desire for fairness. However, their tendency to rate books at the maximum score, regardless of predicted ratings, suggests a more impulsive and emotional approach to evaluating content. This user seems to value diversity and is not afraid to explore unconventional ideas, making them a highly open-minded and adaptable individual. Their love for storytelling and enthusiasm for new experiences may indicate a creative or artistic personality, always seeking inspiration and joy in their reading habits. Overall, this user embodies a unique blend of creativity, curiosity, and practicality, with a penchant for narratives that inspire, educate, and entertain."} \\ \cmidrule{2-2}
    \multicolumn{1}{l}{} &
    \parbox{0.75\textwidth}{item\_list = ["Lust, Money \& Murder - Book 1: A Female Secret Service Agent Takes on an International Criminal - Kindle edition", "Lust, Money \& Murder - Books 1, 2 \& 3: A Female Secret Service Agent Takes on an International Criminal","The Hitman's Lover (Jack Scully Thrillers) (Volume 1)", "Initiation, Human and Solar","Consciousness of the Atom", "Contagious: Why Things Catch On","The Poetical Works of Elizabeth Barrett Browning, Vol. I eBook","The Original Folk and Fairy Tales of Grimm Brothers: The Complete First Edition", "My Short, Happy Life In 'Jeopardy!'", "22/11/63"]\\

    actual\_ratings= [5, 5, 5, 4, 5, 5, 5, 3, 5, 3]\\
    predict\_ratings = [5, 5, 5, 5, 5, 5, 5, 5, 5, 5]} \\ \bottomrule

    \end{tabular}
    }
    \caption{Examples of $\mathcal{S}_2$ and observartions in $\mathcal{W}_3$}
    \label{tab:persona_evolve_3}
    \end{table*}

    \begin{table*}[h!]
      \centering
      \small
      \resizebox{\linewidth}{!}{
      \begin{tabular}{ll}
      \toprule
      \multicolumn{2}{c}{\textbf{Examples of Persona Evolution of \method(\textit{A User from Book Domain})}} \\ \midrule
      \multicolumn{1}{l}{$\mathcal{S}_3$ and observartions in $\mathcal{W}_4$} &
      \parbox{0.75\textwidth}{"The user appears to be a highly enthusiastic and optimistic individual who enjoys a wide range of books across various genres. Their love for storytelling and enthusiasm for new experiences suggests a creative or artistic personality, always seeking inspiration and joy in their reading habits. However, their tendency to rate books at the maximum score, regardless of predicted ratings, suggests a more impulsive and emotional approach to evaluating content.Their consistent 5-star ratings for a variety of titles, including those that blend fiction with spiritual or self-help themes, suggest a deep interest in personal growth and self-discovery. This inclination towards inspirational and motivational content may indicate that the user is seeking guidance or validation in their own life journey. Their appreciation for practical knowledge and skills, as evident in their interest in finance and funding, highlights their pragmatic side and potential entrepreneurial or creative pursuits.However, their 3-star rating for"The Original Folk and Fairy Tales of Grimm Brothers" and 3-star rating for"22/11/63" suggests that they may not always appreciate complex or darker themes, and may prefer more lighthearted or uplifting content. This preference for happier endings and more optimistic themes may be a key aspect of their reading preferences.Their enjoyment of stories that involve justice and problem-solving suggests a strong sense of morality and a desire for fairness. However, their tendency to rate books at the maximum score, regardless of predicted ratings, suggests that they may be more focused on the emotional resonance of a story rather than its literary merit or technical quality.Overall, this user embodies a unique blend of creativity, curiosity, and practicality, with a penchant for narratives that inspire, educate, and entertain. They appear to be a highly open-minded and adaptable individual, always seeking new experiences and inspiration in their reading habits."} \\ \cmidrule{2-2} 
      \multicolumn{1}{l}{} &
      \parbox{0.75\textwidth}{item\_list= [ "Into the Light (Easyread Large Bold Edition): Real Life Stories About Angelic Visits, Visions of the Afterlife, and Other Pre-Death Experiences", "Catnapped: A Klepto Cat Mystery (Volume 1)", "Dancing In The Moonlight (The Cowboys of Cold Creek)", "11/22/63 (Thorndike Press Large Print Core)","The Medium: An Emily Chambers Spirit Medium Novel (Volume 1)","Passion, Power \& Sin - Book 1: The Victim of a Global Internet Scam Plots Her Revenge - Kindle edition", "Passion, Power \& Sin - Books 1-5 (Book 1 Free)", "Season of the Witch","Murder in the South of France: A Maggie Newberry Mystery, Vol. 1 (Maggie Newberry Mysteries)","World hunger: Twelve myths"]\\
  
      actual\_ratings= [5, 4, 5, 3, 5, 5, 5, 5, 4, 5]\\
      predict\_ratings=[5, 5, 5, 3, 5, 5, 5, 5, 5, 4]} \\ \bottomrule
  
      \end{tabular}
      }
      \caption{Examples of $\mathcal{S}_3$ and observartions in $\mathcal{W}_4$}
      \label{tab:persona_evolve_4}
      \end{table*}

    \newpage

    \begin{table*}[h!]
      \centering
      \small
      \resizebox{\linewidth}{!}{
      \begin{tabular}{ll}
      \toprule
      \multicolumn{2}{c}{\textbf{Examples of Persona Evolution of \method(\textit{A User from Book Domain})}} \\ \midrule
     
      \multicolumn{1}{l}{$\mathcal{S}_4$ and observartions in $\mathcal{W}_5$} &
      \parbox{0.75\textwidth}{"Based on the differences between predicted and actual ratings, the user reveals a more nuanced and complex individual. They appear to be a highly enthusiastic and optimistic person, always seeking inspiration and joy in their reading habits. Their love for storytelling and enthusiasm for new experiences suggest a creative or artistic personality, but with a notable emphasis on emotional resonance and personal connection.Their consistent 5-star ratings for a wide range of titles, including those with inspirational and motivational themes, suggest a deep interest in personal growth and self-discovery. However, their ratings also indicate a willingness to engage with more lighthearted and entertaining content, such as romance and mystery novels. This preference for happier endings and more optimistic themes may be a key aspect of their reading preferences.Their enjoyment of stories that involve justice and problem-solving suggests a strong sense of morality and a desire for fairness. However, their tendency to rate books at the maximum score, regardless of predicted ratings, suggests that they may be more focused on the emotional impact of a story rather than its literary merit or technical quality.Interestingly, their ratings also suggest a willingness to engage with complex and thought-provoking topics, such as world hunger and global issues. This inclination towards inspirational and motivational content, combined with a desire for practical knowledge and skills, highlights their pragmatic side and potential entrepreneurial or creative pursuits.They may not always appreciate complex or darker themes, and may prefer more uplifting content. However, their willingness to engage with more serious topics, such as world hunger, suggests a depth and nuance to their personality that is not immediately apparent.Overall, this user embodies a unique blend of creativity, curiosity, and practicality, with a penchant for narratives that inspire, educate, and entertain. They appear to be a highly open-minded and adaptable individual, always seeking new experiences and inspiration in their reading habits. Their refined persona reveals a more complex and multifaceted individual, with a deep appreciation for emotional resonance, personal growth, and practical knowledge."} \\ \cmidrule{2-2}
      \multicolumn{1}{l}{} &
      \parbox{0.75\textwidth}{item\_list= [ "The Quickening of America: Rebuilding Our Nation, Remaking Our Lives", "The Da Vinci Code (Robert Langdon)","The Accidental Cop (Ben Colder Mystery)", "Lingering Echoes", "Rumors (Lingering Echoes)","Murder in Paris (The Maggie Newberry Mystery Series)", "Stilettos \& Scoundrels", "Bitty And The Naked Ladies (The Sherri Travis Mystery Series) eBook", "Sati and the Rider: A Satyana Mystery (Volume 1)","Fleur deKey: a French Quarter Mystery (The Foundation Mystery Series) (Volume 1)"]\\\\
      actual\_ratings= [5, 4, 5, 5, 5, 5, 4, 5, 5, 5]\\
      predict\_ratings=[5, 5, 4, 5, 5, 5, 4, 5, 5, 5]
      } \\ \bottomrule
     
      \end{tabular}
      }
      \caption{Examples of $\mathcal{S}_4$ and observartions in $\mathcal{W}_5$}
      \label{tab:persona_evolve_5}
      \end{table*}
  \label{sec:appendix}
\end{document}